\pdfoutput=1

\documentclass[11pt]{article}

\usepackage[final]{acl}

\usepackage{times}
\usepackage{latexsym}
\usepackage{array} 
\usepackage{booktabs}
\usepackage[most]{tcolorbox}
\usepackage{enumitem}

\usepackage[T1]{fontenc}

\usepackage[utf8]{inputenc}

\usepackage{microtype}

\usepackage{inconsolata}

\usepackage{graphicx}

\title{Somatic in the East, Psychological in the West?: \\ A Clinically-Grounded Evaluation of Cross-Cultural Depression Symptoms in LLMs}

\author{
 \textbf{Shintaro Sakai\textsuperscript{1}},
 \textbf{Jisun An\textsuperscript{1}},
 \textbf{Migyeong Kang\textsuperscript{2}},
 \textbf{Haewoon Kwak\textsuperscript{1}}
\\
\\
 \textsuperscript{1}Indiana University Bloomington, USA,
 \textsuperscript{2}Sungkyunkwan University, Republic of Korea
\\
 \small{
   \textbf{Correspondence:} \href{mailto:email@domain}{shinsaka@iu.edu}
 }
}

\begin{document}
\maketitle
\begin{abstract}
Large language models (LLMs) are increasingly used for mental health applications, raising questions about whether they reflect established clinical knowledge. Clinical psychology has documented systematic cultural differences in the presentation of depression symptoms, with Western populations emphasizing emotional symptoms and many East Asian populations reporting more somatic symptoms.
We evaluate whether general-purpose LLMs reproduce these clinically established cross-cultural patterns using prompts grounded in clinical descriptions of depression. We examine model responses under different cultural personas and languages.
We find that LLMs struggle to reproduce expected cultural patterns when prompted in English. Prompting in major Eastern languages improves alignment in some configurations, suggesting that language cues partially activate cultural knowledge. However, model behavior remains dominated by a strong, culture-invariant hierarchy of depression symptoms that often overrides cultural cues, highlighting limitations of current LLMs for mental health applications.
\end{abstract}

\section{Introduction}
Large language models (LLMs) are increasingly explored for mental health applications, including screening, conversational support, and clinical decision assistance~\cite{hua2025large, obradovichOpportunitiesRisksLarge2024, lawrenceOpportunitiesRisksLarge2024, stade2024large}. Because such systems may influence the interpretation of sensitive mental health signals, their behavior must align with established clinical knowledge. Ensuring that LLMs reason about mental health symptoms in clinically valid ways is therefore essential for their safe deployment.

Clinical psychology has long documented that depression manifests differently across cultural contexts~\cite{dechoudhuryGenderCrossCulturalDifferences2017, loveysCrossculturalDifferencesLanguage2018, rai2025cross}. 
For example, studies using questionnaires~\cite{parkerChineseSomatizeDepression2001, arnaultSomaticDepressiveSymptoms2006, ryderCulturalShapingDepression2008} and clinical interviews~\cite{ryderCulturalShapingDepression2008, biswasCrossCulturalVariations2016} consistently find that individuals in Western settings often emphasize emotional symptoms such as sadness or loss of interest, whereas individuals in many East Asian contexts more frequently report somatic symptoms such as fatigue or physical pain.  
These differences reflect culturally shaped norms of emotional expression and have been consistently observed across decades of clinical research.

As LLMs become increasingly integrated into mental health technologies, an important question arises: \emph{do LLMs reproduce these clinically established cross-cultural patterns of depression symptoms?}
If LLMs disproportionately reflect dominant cultural perspectives while neglecting others~\cite{shah2020predictive}, systems built on top of them may misinterpret symptoms or overlook culturally specific expressions of depression~\cite{abdelkadir2024diverse}.
Despite growing interest in cultural bias and alignment in LLMs, much less attention has been given to whether LLMs reproduce clinically grounded patterns of mental illness documented in psychological research across cultures. As a result, it remains unclear whether these models reason about depression symptoms in ways that align with established clinical findings.

In this work, we introduce a clinically grounded evaluation framework to examine whether general-purpose LLMs reproduce cross-cultural depression symptom patterns  established in clinical psychology. We construct prompts based on clinical descriptions of depressive symptoms and evaluate model responses under different cultural personas and languages. Following prior cross-cultural clinical psychology research, we operationalize cultural personas at the country level (American, Canadian, Australian as Western; Japanese, Chinese, Indian as Eastern)~\cite{parkerChineseSomatizeDepression2001,
arnaultSomaticDepressiveSymptoms2006, ryderCulturalShapingDepression2008, biswasCrossCulturalVariations2016}. Symptom expression is assessed using a predefined set of DSM-5–based depression symptoms~\cite{ryderCulturalShapingDepression2008}. This setup enables systematic comparison between LLM-generated symptom profiles and  established clinical baselines.

Our results reveal several limitations in the cultural reasoning abilities of current LLMs. When prompted in English, LLMs struggle to reproduce the expected cultural patterns of depression symptoms. Prompting in major Eastern languages improves alignment in several configurations, suggesting that language cues partially activate culturally grounded knowledge. However, deeper analysis reveals two key challenges: weak sensitivity to cultural personas and a strong, culture-invariant hierarchy of depression symptoms that often overrides cultural cues. These findings suggest that current LLMs lack the robust cultural reasoning required for reliable mental health applications.

\section{Background}
\label{sec:background}
\subsection{LLM Applications in Mental Health Contexts}
\looseness=-1 LLMs are increasingly used in mental health. A recent review shows a surge in related publications in 2023~\cite{hua2025large}. 
Their applications fall into three broad areas~\cite{hua2025large}:
1) conversational agents for digital companionship and emotional support~\cite{hu2024psycollm, laiPsyLLMScalingGlobal2023, maUnderstandingBenefitsChallenges2024, suharwardy2023feasibility, leeChainEmpathyEnhancing2024, zhangAskExpertLeveraging2023, kumarExploringDesignPrompts2022}, 2) resource enrichment, such as generating synthetic data and educational materials~\cite{yangInterpretableMentalHealth2023, kumarExploringUseLarge2023, mentalllama}, and 3) classification tasks for conditions like depression severity and suicide risk~\cite{yangInterpretableMentalHealth2023,mentalllama, xuMentalLLMLeveragingLarge2024, lamichhaneEvaluationChatGPTNLPbased2023, qi2025supervised, nguyen2024leveraging}.
These use cases demonstrate the expanding role of LLMs in mental health contexts.

\subsection{Cultural Differences in Depression Symptom Expressions}
In clinical psychology, numerous studies support that individuals from Western cultures tend to emphasize psychological symptoms, while those from Eastern cultures tend to emphasize somatic symptoms \cite{kleinman1982neurasthenia, tsoi1985mental, ryderCulturalShapingDepression2008, arnaultSomaticDepressiveSymptoms2006, parker2005depression, juckett2010recognizing, biswasCrossCulturalVariations2016, dere2013beyond, kirmayer2016culture}.

For instance, a series of studies \cite{ryderCulturalShapingDepression2008,dere2013beyond} found that Chinese patients consistently reported more somatic symptoms in interviews and problem reports compared to their Euro-Canadian counterparts, who emphasized psychological symptoms.
Another study focused on depression in Japanese and American college women, finding that Japanese participants reported higher overall somatic distress~\cite{arnaultSomaticDepressiveSymptoms2006}. Similarly, Parker et al. found that 60\% of Malaysian Chinese patients presented with somatic symptoms of depression, compared to only 13\% of Australian Caucasians. While Chinese patients scored higher on somatic items in an inventory, they were less likely to acknowledge psychological symptoms \cite{parkerChineseSomatizeDepression2001}. 

Diagnostic practices also reflect this divide, with Indian psychiatrists prioritizing somatic symptoms (e.g., pain, sleep issues) and American psychiatrists prioritizing cognitive and emotional ones (e.g., pessimism about the future) \cite{biswasCrossCulturalVariations2016}.

These studies highlight cultural differences in how depression is expressed: Eastern populations tend to show somatic symptoms, while Western populations emphasize psychological ones. 
The most widely accepted explanation for somatization among Eastern populations is that it offers a socially safer way to express mental health problems in cultures where mental illness is highly stigmatized. By framing distress in physical terms,  individuals can seek support without being labeled mentally ill \cite{link1997stigma, goldberg1988somatic, barney2006stigma, kung2008symptom, juckett2010recognizing}.

\subsection{Sociocultural Limitations in LLMs}
\subsubsection{Cultural Bias in LLMs}
Several studies have demonstrated that cultural bias exists across different models, typically using the personas method to quantify cultural bias in LLMs~\cite{santy2023nlpositionality, caoAssessingCrossCulturalAlignment2023, alkhamissiInvestigatingCulturalAlignment2024, kharchenkoHowWellLLMs2024, rao2025normad}.

It is well known that the cultural values reflected in LLMs tend to align more closely with the values of the U.S. and other English-speaking countries \cite{johnson2022ghost, santy2023nlpositionality, caoAssessingCrossCulturalAlignment2023, alkhamissiInvestigatingCulturalAlignment2024, rao2025normad}. Prompting LLMs in a country’s local language has been shown to improve cultural alignment \cite{linFewshotLearningMultilingual2022, caoAssessingCrossCulturalAlignment2023, alkhamissiInvestigatingCulturalAlignment2024}. 

These studies often use sociological benchmarks such as the World Values Survey\footnote{https://www.worldvaluessurvey.org/wvs.jsp} and Hofstede’s cultural dimensions \cite{hofstede2001culture} to assess cultural alignment. While these studies provide valuable insights into general cultural value alignment, relying on broad sociological benchmarks may not capture nuanced, domain-specific cultural variations such as those in mental health symptom reporting. This gap is particularly salient in depression, where cultural differences in symptom expression are well-documented in clinical literature but remain underexplored in LLM behavior. 

\subsubsection{Demographic and Diagnostic Biases in Mental Health}
Several studies have examined biases in LLMs within mental health contexts. One study focusing on Borderline Personality Disorder (BPD) and Narcissistic Personality Disorder (NPD) found that GPT-3.5 and GPT-4 exhibited gender bias in diagnostic assessments, particularly against women \cite{chansiriAddressingGenderBias2024}. Another study investigated classification performance of 10 different LLMs across various demographic factors. While models generally performed well with respect to gender and age, their performance varied when factors such as religion and nationality were considered \cite{wang2024unveiling}. The study by~\citet{bouguettaya2025racial} evaluates four large language models on psychiatric vignettes written in race-neutral, race-implied, and race-explicit versions. It reveals that while diagnoses stay mostly consistent across race conditions, treatment recommendations often shift when the patient is described as African American, revealing significant racial bias.

These findings highlight the importance of evaluating LLM bias in mental health applications, given the growing use of LLMs in this domain, and also the scarcity of research addressing cultural bias in such contexts.

\section{Research Hypotheses}

Building upon the background reviewed in \S\ref{sec:background}, we investigate whether LLMs select depression symptoms in ways consistent with cultural patterns identified in clinical psychology.

\begin{itemize}[label={}]
    \item \textbf{H1.} LLMs select psychological symptoms more often for Western cultural personas, and somatic symptoms more often for Eastern cultural personas.
    \item \textbf{H2.} Prompts written in the local language of a country increase cultural alignment in symptom selection.
\end{itemize}

\section{Task Design for Hypothesis Testing}

We assign the model a cultural persona with depression (e.g., \emph{an American person with depression}) and prompt it to select symptoms from a predefined list of 14 depression symptoms (see Table~\ref{tab:symptom_combined_avg}).
These symptoms are extracted from the PsySym dataset~\cite{zhang2022symptom} which is based on the DSM-5 (Diagnostic and Statistical Manual of Mental Disorders, Fifth Edition)~\cite{american2013diagnostic}. Following prior clinical research~\cite{ryderCulturalShapingDepression2008}, these symptoms are categorized as either somatic (e.g., fatigue, sleep disturbance) or psychological (e.g., depressed mood, worthlessness) symptoms.

We use two prompt forms: an implicit culture prompt (ICP) providing only general cultural context, and an explicit culture prompt (ECP) that directly instructs the model to consider the persona's cultural background, allowing us to assess the effect of explicit instruction. To ensure LLMs correctly understand each symptom defined in DSM-5, we provided their brief descriptions in the prompts. These descriptions were derived from the PsySym dataset~\cite{zhang2022symptom}. Full prompts are in \S\ref{sec:prompt} of the Appendix.





While we refer to cultural personas, we operationalize them at the country level, using national identity as a proxy for broader cultural context. This follows conventions in clinical psychology research~\cite{parkerChineseSomatizeDepression2001,
arnaultSomaticDepressiveSymptoms2006, ryderCulturalShapingDepression2008, biswasCrossCulturalVariations2016} and cross-cultural NLP research~\cite{caoAssessingCrossCulturalAlignment2023, abdelkadir2024diverse, alkhamissiInvestigatingCulturalAlignment2024, kharchenkoHowWellLLMs2024, rai2025cross}, where cultural groupings are often defined nationally. We acknowledge that this simplification may overlook within-region and -country variation, but we adopt it for comparability with prior clinical research and to maintain experimental clarity. Specifically, we use Japanese, Chinese, and Indian to represent Eastern cultural groups, and American, Canadian, and Australian to represent Western cultural groups, focusing on countries commonly examined in previous clinical psychology studies~\cite{parkerChineseSomatizeDepression2001,
arnaultSomaticDepressiveSymptoms2006, ryderCulturalShapingDepression2008, biswasCrossCulturalVariations2016}. 

Formally, let 
$S_{\text{som}}$ and
$S_{\text{psy}}$ denote the sets of somatic and psychological symptoms. Let    
$\mathcal{C}_W$ and 
$\mathcal{C}_E$ represent the set of Western and Eastern cultural groups, and let \( x \in \{\text{I}, \text{E}\} \) indicate the prompt type, either implicit (ICP) or explicit (ECP) culture prompt. Then, for each cultural group $c$, we denote the probability that the model selects symptom $s$ as \(P(s | \textsc{P}_x^c)\) where $\textsc{P}_x^c$ is the prompt corresponding to cultural persona $c$ under prompt type $x$.

With \(g \in \{\text{Western}, \text{Eastern}\} \), we compute the group-level selection probability for each symptom as: 
\[
P(s \mid \textsc{p}_x^g) = \frac{1}{|\mathcal{C}_g|} \sum_{c \in \mathcal{C}_g} P(s \mid \textsc{p}_x^c)
\]

The sum of selection probabilities for symptom types is then: 
\begin{align*}
P(\text{somatic} \mid \textsc{p}_x^g) &= \sum_{s \in S_{\text{som}}} P(s \mid \textsc{p}_x^g) \\
P(\text{psychological} \mid \textsc{p}_x^g) &= \sum_{s \in S_{\text{psy}}} P(s \mid \textsc{p}_x^g)
\end{align*}

We define the cultural alignment as the difference in somatic selection:
\[
_1\mathcal{A}_x = P(\text{somatic} \mid \textsc{p}_x^{\text{Eastern}}) - P(\text{somatic} \mid \textsc{p}_x^{\text{Western}})
\]
, which is equivalent to \( P(\text{psychological} \mid \textsc{p}_x^{\text{Western}}) - P(\text{psychological} \mid \textsc{p}_x^{\text{Eastern}}) \), given that symptoms are exhaustively categorized as somatic or psychological. 
A positive $_1\mathcal{A}_x$ indicates that models follow clinical findings by selecting more somatic symptoms for Eastern personas and more psychological symptoms for Western personas. 

Following recent studies showing that prompts written in non-English languages can elicit responses that are more culturally aligned with the culture of the language \cite{linFewshotLearningMultilingual2022, caoAssessingCrossCulturalAlignment2023,alkhamissiInvestigatingCulturalAlignment2024}, we test prompts written in the local language spoken in each Eastern cultural group. Specifically, we use Japanese for Japanese, Chinese for Chinese, and Hindi for Indian. While we acknowledge that multiple languages and dialects are spoken within each country, we selected the most widely spoken language in each country. 
Native speakers validated the quality of the prompts for each language.
For simplicity, we refer to English prompts as the {English Language Prompt (ENG-P)}, and those written in local language as the {Local Language Prompt (LOC-P)}. Let \( l(c) \) be a function that maps each Eastern cultural group \( c \in \mathcal{C}_\text{Eastern}\) to its local language (e.g., \( l(\text{Japan}) = \text{Japanese} \)). We denote $_1\mathcal{A}_x(l(c))$ as the cultural alignment tested by LOC-P for Eastern cultural group $c$, and $_1\mathcal{A}_x(Eng)$ as that by ENG-P. 
\begin{align*}
_1\mathcal{A}_x(l(c)) &= P(\text{somatic} \mid \textsc{p}_x^{\text{Eastern}}, l(c)) \\
&\quad - P(\text{somatic} \mid \textsc{p}_x^{\text{Western}}, \text{English})
\end{align*}

This formulation allows us to test the hypotheses as:

\smallskip

\noindent H1 is supported when $_1\mathcal{A}_x > 0$, and \\H2 is supported when $_1\mathcal{A}_x(l(c)) > {_1}\mathcal{A}_x(Eng)$.

\smallskip

Each model is tested across three symptom-choice conditions (LLMs can select one, three, or five symptoms per iteration), two prompt types (I/E), and six countries (three Western, three Eastern), yielding 36 configurations under ENG-P and 18 under LOC-P. We run 100 iterations per setting to ensure reliable results.

Throughout the paper, we use both symbolic notation and plain language interchangeably to improve readability and ease of understanding.

\subsection{Six Language Models for Evaluation}
We evaluate a total of six LLMs. We select four open-source models for their accessibility and replicability: Llama-3.1-8B-it \cite{touvronLlamaOpenEfficient2023},  Gemma-7B-it \cite{teamGemmaOpenModels2024}, Qwen-2.5-7B-it \cite{baiQwenTechnicalReport2023}, and DeepSeek-R1-Distill-Qwen-7B \cite{deepseek-aiDeepSeekV3TechnicalReport2025}. 
We additionally include GPT-4o~\cite{openaiGPT4TechnicalReport2024} as a widely adopted proprietary baseline. We also test MentaLLaMA-chat-7B~\cite{mentalllama}, an open-source LLM fine-tuned on the IMHI mental health dataset.
In our preliminary experiments, we tested temperatures of 0.5, 0.7, and 1.0, but observed no significant differences. Thus, we report results generated with a temperature of 0.7, a standard setting in prior work~\cite{chansiriAddressingGenderBias2024, alkhamissiInvestigatingCulturalAlignment2024}.

\section{Results}


\subsection{Cultural Alignment in English Prompts (H1)}

\begin{figure*}[h] 
    \centering
    \includegraphics[width=1.0\textwidth]{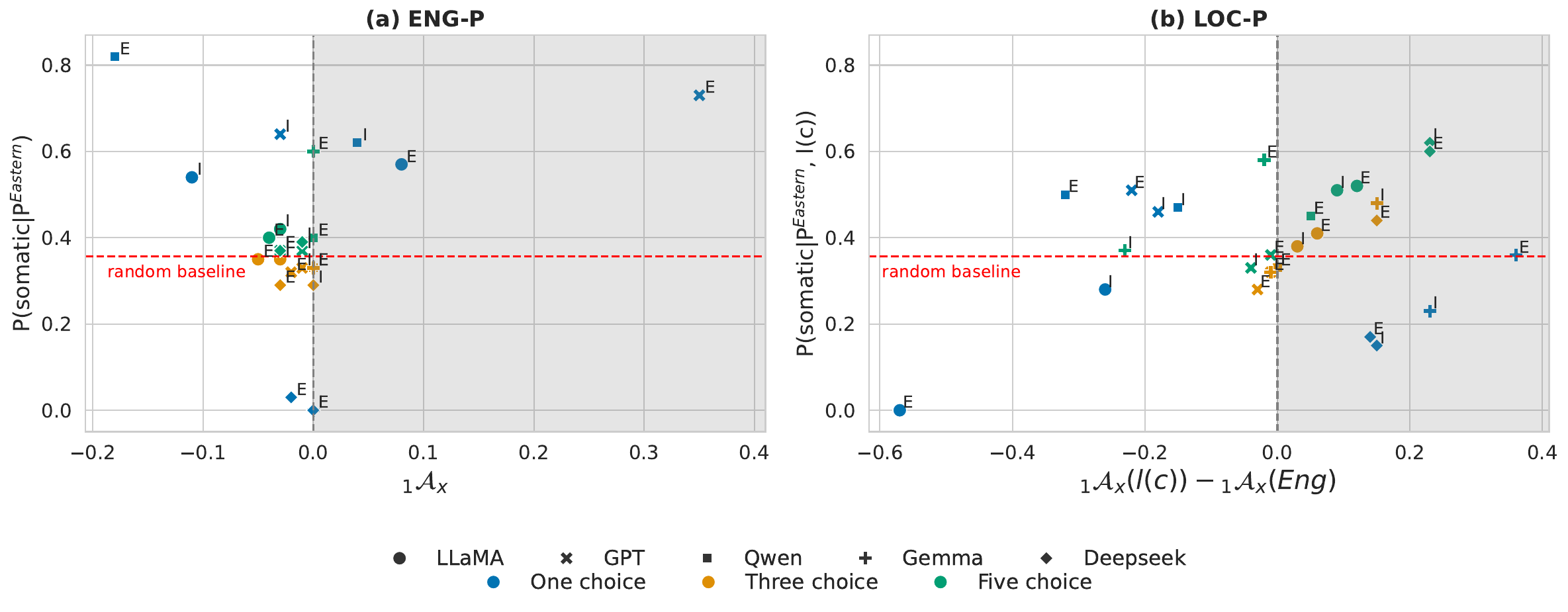} 
    \caption{In (a), the $x$-axis shows cultural alignment $_1\mathcal{A}_x$ under the ENG-P condition; values greater than 0 indicate alignment with prior clinical psychology findings. In (b), values greater than 0 on the $x$-axis indicate an \emph{increase} in alignment under the LOC-P condition. Higher values on the $y$-axis reflect a stronger tendency for the model to choose \( S_{\text{som}} \). \emph{I} and \emph{E} indicate implicit (ICP, $\textsc{p}_{I}$) and explicit culture prompt (ECP, $\textsc{p}_{E}$), respectively. Culturally aligned regions are shaded to help readers visually identify expected model behavior.}
    \label{fig:task1_alignment} 
\end{figure*}

Figure~\ref{fig:task1_alignment}(a) shows each model's alignment level under the ENG-P condition, for both \(\textsc{p}_{I} \) and \(\textsc{p}_{E} \). 
The $x$-axis presents cultural alignment $_1\mathcal{A}$, defined as \( P(\text{somatic} \mid \textsc{p}_x^{\text{Eastern}}) - P(\text{somatic} \mid \textsc{p}_x^{\text{Western}}) \). 
Positive values of $_1\mathcal{A}$ indicate alignment with prior clinical psychology findings that somatic symptoms are more common in Eastern contexts. The $y$-axis shows \(P(\text{somatic} \mid \textsc{p}^{\text{Eastern}})\), the average proportion of \( S_{\text{som}} \) selected by Eastern personas. Since 5 out of 14 symptoms are somatic, the random baseline is 5/14 ($\approx$0.357); values above this indicate a bias toward somatic symptom selection. We use the $y$-axis to examine whether higher absolute somatic bias corresponds to larger cultural gaps in $_1\mathcal{A}$. 
Although we attempted to evaluate the mental health-specific models such as MentaLLaMA~\cite{mentalllama}, they consistently failed to generate valid outputs due to safety-alignment constraints (Please see \S\ref{sec:mental-llama} of the Appendix). We therefore excluded MentaLLaMA from the rest of the analyses.

Overall, LLMs' behaviors do not align with expectations under the ENG-P condition. Of the 30 tested settings (5 models $\times$ 3 choice conditions $\times$ 2 prompt types), only three settings---Llama (one-choice, \(\textsc{p}_{E} \)), GPT (one-choice, \(\textsc{p}_{E} \)), and Qwen (one-choice, \(\textsc{p}_{I} \))---show patterns consistent with prior clinical findings. In general, absolute values of $_1\mathcal{A}$ are small.
Differences are particularly marginal in the three- and five-choice conditions, where absolute values of $_1\mathcal{A}$ typically range from 0.03 to 0.05. This suggests that assigning Western or Eastern cultural personas has limited influence on LLM symptom selection in multi-choice conditions. A full breakdown of symptom selection rates across all model and prompt configurations is available in Table \ref{tab:symptom-region-model-all-en-task1} in \S\ref{sec:alignment} of the Appendix\footnote{While we aim to make the main text self-contained, the number and complexity of experimental conditions make it impractical to include all results. To support transparency and reproducibility, we provide detailed results in the Appendix.}.

Llama, GPT, and Qwen exhibit a somatic symptom selection bias in the one-choice condition, primarily driven by a strong preference for s2 (Decreased energy, tiredness, and fatigue; Somatic). In contrast, DeepSeek shows a psychological symptom selection bias, largely due to a strong preference for s3 (Depressed mood; Psychological) (See \S\ref{sec:overall} in the Appendix). 
Higher absolute somatic bias, captured by \(P(\text{somatic} \mid \textsc{p}^{\text{Eastern}})\), does not correspond to larger cultural gaps in alignment ($_1\mathcal{A}$).

In summary, \emph{H1 is not supported}, as the majority of experimental conditions fail to align with findings from prior clinical psychology research.

\subsection{Effect of Language on Alignment (H2)} 
\label{sec:languageeffect}
Interestingly, of the 30 settings, 15 exhibit increased alignment under the LOC-P (Figure~\ref{fig:task1_alignment}(b)). Positive values of $_1\mathcal{A}_x(l(c)) - {_1}\mathcal{A}_x(Eng)$ on the $x$-axis indicate the alignment increase. Llama shows improvement in the three- and five-choice conditions, Qwen in the five-choice condition, and Gemma primarily in the one-choice condition. DeepSeek demonstrates consistent alignment across all conditions, whereas GPT shows decreased alignment throughout. These results suggest that prompt language can affect alignment levels in some models. Detailed results are provided in Table \ref{tab:symptom-region-model-all-dif} and \ref{tab:x_diff_table} in \S\ref{sec:alignment} of the Appendix.

\begin{table*}[h]
    \centering
    \resizebox{\textwidth}{!}{%
    \begin{tabular}{lrrrrrrrrrrr}
        \toprule
        & \textbf{All} & \textbf{Llama} & \textbf{GPT} & \textbf{Qwen} & \textbf{Gemma} & \textbf{DeepSeek} & \textbf{One Choice} & \textbf{Three Choice} & \textbf{Five Choice} & \textbf{\(\textsc{p}_{I} \)} & \textbf{\(\textsc{p}_{E} \)} \\
        \midrule
        t-stat & -0.13 & 0.79 & 2.15 & 1.03 & -0.93 & \textbf{-10.02} & 0.89 & -2.10 & -1.09 & -0.34 & 0.08\\
        p-value & 0.90 & 0.46 & 0.09 & 0.35 & 0.39 & \textbf{0.00} & 0.39 & 0.07 & 0.30 & 0.74 & 0.94\\
        \bottomrule
    \end{tabular}
    }
    \caption{Paired t-test by model, experimental condition, and prompt type. Statistically significant values (p $<$ 0.05) are bolded.}
    \label{tab:paired_ttest_task1}
\end{table*}

To assess the impact of language on alignment, we conducted paired $t$-tests, as in Table~\ref{tab:paired_ttest_task1}. 
Each $t$-test evaluates alignment within a distinct experimental condition (e.g., model, choice condition, or prompt type). As our goal is not to test a single global hypothesis, but rather to probe how alignment shifts across various independent conditions, we do not apply multiple comparisons correction. 
Overall, alignment increases slightly (t$=$-0.13) but is not statistically significant (p$=$0.90). At the model level, both DeepSeek and Gemma show improved alignment, with statistical significance observed only for DeepSeek (p$<$0.0005). By choice condition, both the three- and five-choice conditions show increases, with a larger effect in the three-choice. However, neither reached statistical significance (p$=$0.07 and 0.30, respectively). At the prompt level, \(\textsc{p}_{I} \) also shows a modest, non-significant increase (p$=$0.74). These results suggest that not all models are equally sensitive to language-induced cultural cues, and the effectiveness of prompt language depends on the choice condition and prompt type. 

In summary, the \emph{results do not support H2} overall, though DeepSeek shows a statistically significant increase in alignment, providing model-level support.


\subsection{Sensitivity to Cultural Personas}
\label{sec:sensitivityculturalpersonas}
To quantify the effect of Western and Eastern personas on symptom selection, we examine each model's \emph{persona sensitivity}.
Persona sensitivity refers to how well a model differentiates symptom selection patterns between Western and Eastern cultural personas, measured by the cosine similarity between the distributions \(P(s \mid \textsc{p}_x^{\text{Eastern}})\) and \(P(s \mid \textsc{p}_x^{\text{Western}})\) within the same prompt type. Lower cosine similarity values indicate higher sensitivity.

Overall, models exhibit significantly low persona sensitivity under the ENG-P condition, with cosine similarity values ranging from 0.77 to 1.00 (Table~\ref{tab:sensitivity}). Specifically, sensitivity is low in the three- and five-choice settings (e.g., cosine similarity $\geq$ 0.95), suggesting that cultural distinctions weaken as the number of selectable symptoms increases. For Llama and GPT, persona sensitivity improves from \(\textsc{p}_{I} \) to \(\textsc{p}_{E} \) (e.g., Llama: 0.98 $\rightarrow$ 0.95, GPT: 0.99 $\rightarrow$ 0.77 in one-choice), indicating that explicitly prompting for cultural consideration helps these models better distinguish between Western and Eastern cultural personas. However, the effectiveness of \(\textsc{p}_{E} \) remains limited as the effect is not observed in Qwen or DeepSeek. Under the LOC-P condition, persona sensitivity increases in 28 out of 30 experimental conditions (3 choice conditions $\times$ 2 prompt types $\times$ 5 models). As with the ENG-P condition, the impact of \(\textsc{p}_{E} \) is confined to specific experimental configurations. 

Importantly, low persona sensitivity under the ENG-P condition suggests that internal model tendencies override the influence of cultural personas or prompt variations. Although the findings indicate that prompting in local languages leads to more distinct symptom selection behaviors between Western and Eastern personas, higher sensitivity does not necessarily imply better cultural alignment. For example, while Llama shows greater sensitivity under the LOC-P, this does not translate to improved alignment with clinically observed patterns.

\begin{table*}[t]
\centering
\resizebox{\textwidth}{!}{%
\begin{tabular}{lccccccccccc}
\toprule
 & \multicolumn{2}{c}{Llama} & \multicolumn{2}{c}{GPT} & \multicolumn{2}{c}{Qwen} & \multicolumn{2}{c}{Gemma} & \multicolumn{2}{c}{DeepSeek} \\
& ENG-P & LOC-P & ENG-P & LOC-P & ENG-P & LOC-P & ENG-P & LOC-P & ENG-P & LOC-P \\
\midrule
One choice   & 0.98/0.95 & 0.31/0.51 & 0.99/0.77 & 0.91/0.92  & 0.99/0.98 & 0.63/0.83 & 0.89/0.89 & 0.84/0.00 & 0.95/0.99 & 0.39/0.32 \\
Three choice & 0.99/0.98 & 0.67/0.81 & 1.00/0.95 & 0.99/0.99 & 1.00/1.00 & 0.96/0.83 & 1.00/1.00   & 0.95/0.97   & 1.00/1.00  & 0.95/0.94 \\
Five choice  & 0.98/0.98 & 0.84/0.85 & 1.00/1.00 & 0.96/0.98 & 1.00/1.00 & 0.88/0.83 & 1.00/1.00 & 0.81/0.82 & 1.00/0.99 & 0.86/0.84 \\
\bottomrule
\end{tabular}
}
\caption{Cosine similarities between \(P(s \mid \textsc{p}_x^{\text{Eastern}})\) and \(P(s \mid \textsc{p}_x^{\text{Western}})\) distributions under the ENG-P and LOC-P conditions. Two cosine similarity values under each prompt correspond to the cosine values for \(\textsc{p}_{I} \) and \(\textsc{p}_{E} \).}
\label{tab:sensitivity}
\end{table*}

\subsection{Symptom Preference Hierarchy}
Persona sensitivity analysis reveals that models tend to select similar symptoms for Western and Eastern cultural personas, particularly under the ENG-P condition.
To further examine this universal symptom preference, we averaged \(P(s | \textsc{P}_x^c)\) across 180 experimental settings (6 countries $\times$ 3 choice conditions $\times$ 2 prompt types $\times$ 5 models) for ENG-P and 90 experimental settings (3 countries $\times$ 3 choice conditions $\times$ 2 prompt types $\times$ 5 different models) for LOC-P. 
Before averaging, \(P(s | \textsc{P}_x^c)\) values were normalized by the maximum possible selection rate in each choice condition: 1/3 for three-choice and 1/5 for five-choice. 
For example, in the three-choice setting (3 choices $\times$ 100 iterations = 300 total selections), a symptom chosen in all iterations would have a maximum selection rate of 100/300 = 1/3.

Table \ref{tab:symptom_combined_avg} shows that under the ENG-P condition, LLMs consistently favor certain symptoms, particularly s2 (Decreased energy, tiredness, and fatigue; Somatic), s3 (Depressed mood; Psychological), and s8 (Loss of interest or motivation; Psychological) with average selection rates of 0.6 for s2 and s3, and 0.4 for s8. In contrast, symptoms like s12 (Suicidal ideas; Psychological), s10 (Poor memory; Psychological), and s7 (Indecisiveness; Psychological) are rarely chosen (0.00–0.02). A similar trend appears under the LOC-P condition (Table \ref{tab:symptom_combined_avg}), where s2 and s3 remain the most selected ($\simeq$ 0.6). While s8 is still frequently selected at 0.27, s1 (Anger and irritability; Psychological) rises to 0.41, becoming the third most selected. s7, s10 and s12 show slight increases to 0.06 but remain rarely selected, along with s6 (Inattention; Psychological) and s9 (Pessimism; Psychological).

These results suggest that LLMs have a hierarchical understanding of depression symptoms, consistently identifying some as more representative. This likely reflects the frequency of each symptom in training data. Importantly, the models' preference patterns often outweigh the influence of cultural personas. The underrepresentation of s12 may also stem from safety mechanisms that suppress suicide-related content~\cite{li2025safety}, potentially limiting the models’ ability to identify critical symptoms.


\begin{table}[h]
\centering
\resizebox{\columnwidth}{!}{%
\begin{tabular}{p{7.5cm}cc}
\toprule
Symptom Name & ENG-P & LOC-P \\
\midrule
s1: Anger and irritability (Psychological) & 0.22 & \textbf{0.41} \\
s2: Decreased energy, tiredness (Somatic) & \textbf{0.62} & \textbf{0.64} \\
s3: Depressed mood (Psychological) & \textbf{0.59} & \textbf{0.60} \\
s4: Genitourinary symptoms (Somatic) & 0.07 & 0.14 \\
s5: Hyperactivity and agitation (Somatic) & 0.10 & 0.14 \\
s6: Inattention (Psychological) & 0.08 & 0.06 \\
s7: Indecisiveness (Psychological) & 0.02 & 0.06 \\
s8: Loss of interest or motivation (Psych.) & \textbf{0.39} & 0.27 \\
s9: Pessimism (Psychological) & 0.10 & 0.06 \\
s10: Poor memory (Psychological) & 0.01 & 0.04 \\
s11: Sleep disturbance (Somatic) & 0.15 & 0.20 \\
s12: Suicidal ideas (Psychological) & 0.00 & 0.06 \\
s13: Weight/appetite change (Somatic) & 0.06 & 0.11 \\
s14: Worthlessness and guilt (Psych.) & 0.09 & 0.10 \\
\bottomrule
\end{tabular}
}
\caption{Average symptom proportions under ENG-P and LOC-P conditions.}
\label{tab:symptom_combined_avg}
\end{table}

\subsection{Baseline Analysis}
We also examine whether LLMs' symptom selection behavior under the non-cultural baseline aligns more closely with Western or Eastern persona settings under the ENG-P condition. In the non-cultural baseline, we remove cultural labels from the prompt (e.g., ``person with depression'').

Overall, symptom distributions under the non-cultural baseline closely resemble those of both Western and Eastern personas across most models and conditions (Table~\ref{tab:sensitivity-base}). Importantly, there is no consistent evidence that the baseline aligns more strongly with either group. In the three- and five-choice settings, the baseline becomes nearly identical to both personas, indicating that none of the personas, including the non-cultural baseline meaningfully diverge when models can select a broader range of symptoms.

Although our earlier persona sensitivity analysis shows that the one-choice condition produces the strongest persona effects, even this setting reveals only marginal differences between the baseline and Western/Eastern personas. The only exception is Gemma under \(\textsc{p}_{E} \), where cosine similarities to both Western and Eastern distributions are zero due to deterministic symptom selection (see Figure~\ref{fig:combined-heatmaps-one} and ~\ref{fig:combined-heatmaps-one-eastern} in Appendix). In other models, baseline–Western and baseline–Eastern gaps are somewhat larger in the one-choice setting (e.g., Llama: Western = 0.97 vs. Eastern = 0.94 under \(\textsc{p}_{I} \); Western = 0.83 vs. Eastern = 0.94 under \(\textsc{p}_{E} \)), but the direction and magnitude of these differences are inconsistent across models and prompts. Overall, the non-cultural baseline is neither clearly Western-coded nor Eastern-coded.

\begin{table*}[t]
\centering
\resizebox{\textwidth}{!}{%
\begin{tabular}{lccccccccccc}
\toprule
 & \multicolumn{2}{c}{Llama} 
 & \multicolumn{2}{c}{GPT} 
 & \multicolumn{2}{c}{Qwen} 
 & \multicolumn{2}{c}{Gemma} 
 & \multicolumn{2}{c}{DeepSeek} \\
\cmidrule(lr){2-3}
\cmidrule(lr){4-5}
\cmidrule(lr){6-7}
\cmidrule(lr){8-9}
\cmidrule(lr){10-11}
& Western & Eastern & Western & Eastern & Western & Eastern & Western & Eastern & Western & Eastern \\
\midrule
One choice   & 0.97/0.83 & 0.94/0.94 & 0.98/0.88 & 0.99/0.97  & 0.81/1.00 & 0.85/0.98 & 0.90/0.00 & 1.00/0.00 & 0.93/0.82 & 0.94/0.81 \\
Three choice & 0.99/0.94 & 0.98/0.96 & 1.00/1.00 & 1.00/0.95 & 0.97/0.95 & 0.97/0.95 & 1.00/1.00   & 1.00/1.00   & 1.00/0.98  & 1.00/0.99 \\
Five choice  & 0.99/0.95 & 0.98/0.97 & 1.00/1.00 & 1.00/1.00 & 1.00/1.00 & 1.00/1.00 & 1.00/1.00 & 1.00/1.00 & 1.00/1.00 & 1.00/0.98 \\
\bottomrule
\end{tabular}
}
\caption{Cosine similarities between \(P(s \mid \textsc{p}_x^{\text{Baseline}})\) and \(P(s \mid\textsc{p}_x^{\text{Western}})\), and between \(P(s \mid \textsc{p}_x^{\text{Baseline}})\) and \(P(s \mid \textsc{p}_x^{\text{Eastern}})\) distributions under the ENG-P condition. Two cosine similarity values under each prompt correspond to the cosine values for \(\textsc{p}_{I} \) and \(\textsc{p}_{E} \).}
\label{tab:sensitivity-base}
\end{table*}

\section{Discussion}
Overall, our findings reveal substantial inconsistencies between LLM outputs and widely-recognized clinical patterns of depression symptom expression across cultures. Beyond simply indicating model underperformance, these results highlight several critical challenges in developing culturally aligned mental health AI.

First, the misalignment suggests that general-purpose training data is insufficient for capturing clinical nuance. LLMs trained on large-scale web corpora likely lack the foundational knowledge required to replicate culturally sensitive reasoning grounded in clinical psychology. 

Second, our results suggest that the DSM-5, developed within Western clinical contexts, may not fully capture culturally specific manifestations of depression prevalent in Eastern or non-Western populations~\cite{ecksStrangeAbsenceThings2016}. This limitation could skew LLM behavior by shaping what is considered ``depression'' in our evaluation. 

Third, the East–West distinction in symptom expression may be more situational than stable. Prior research suggests that this cultural pattern is influenced by the mode of assessment (e.g., interviews vs. questionnaires)~\cite{simon1999international, yeung2004illness, ryderCulturalShapingDepression2008}. If LLMs primarily reflect textual discourse, they may miss cultural patterns that emerge in other settings. 

Lastly, the ENG-P condition may reflect how Eastern individuals with mental disorders are portrayed through Western perspectives in English-language texts. This suggests that the model's output might be capturing a ``Western perspective'' rather than Eastern self-expression.

Our task design has important real-world implications. For example, AI-powered mental health chatbots or diagnostic tools could be adapted to emphasize somatic symptoms for users from Eastern countries and psychological symptoms for those from Western countries. Such culturally informed adjustments may improve interactions and diagnostic accuracy. However, our findings suggest that current general-purpose LLMs are unlikely to make these distinctions when cultural identities are introduced solely through prompt-based personas.

Furthermore, the models' lack of nuanced cultural reasoning was not only evident in the symptom selection task but was also confirmed in the inverse cultural attribution task (\S\ref{sec:culturalattribution} in Appendix). There, models were given a symptom and asked to choose which of two cultural personas (i.e., one Western and one Eastern) was more likely to express it. A further fine-grained symptom-level analysis of psychological symptoms (\S\ref{sec:individual}) similarly shows limited alignment with clinically observed patterns. Our analysis also uncovered other concerning behaviors, such as a high degree of determinism in some models and various idiosyncratic biases, which are detailed in the Appendix (\S\ref{sec:determinismsymptom} and \S\ref{sec:symptomselectiondifferences}).


\section{Conclusion}
To the best of our knowledge, this is the first study to examine whether large language models reproduce clinically established cross-cultural patterns of depression symptoms. 
We view this work as a foundational step toward understanding how general-purpose LLMs associate depression symptoms with culture. To explore this question, we evaluated multiple widely used LLMs across a range of experimental configurations, including different prompt formats, cultural personas, prompt variations, and languages.

Our findings reveal several limitations in the cultural reasoning abilities of current LLMs. When prompted in English, the models struggle to reproduce clinically established cross-cultural patterns of depression symptoms. Prompting in local languages partially improves cultural alignment in symptom selection, suggesting that language cues can activate culturally grounded knowledge. However, model responses remain dominated by a strong, culture-invariant hierarchy of depression symptoms that often overrides cultural cues.

These results raise important concerns about the reliability of general-purpose LLMs in culturally grounded mental health contexts and highlight the need for more robust approaches to incorporating culturally sensitive clinical knowledge into LLM development and evaluation.

\clearpage

\section*{Limitations}
\label{sec:limitation}
There are several limitations in this study. 
One limitation is the simplification of Eastern vs. Western categorization. Symptom expression can vary not only between regions but also within regions, countries, and individuals. While differences between countries in the same region such as China and Japan likely exist, prior clinical psychology studies compare one Eastern country with one Western country, making it difficult to analyze intra-regional variation. While within-country or individual-level differences may be present, existing literature generally treats the country as the primary unit of analysis. Following this convention allows for direct comparison with prior work; therefore, such finer-grained differences are beyond the scope of the current study.

Secondly, our study is limited by its use of a broad category of somatic symptoms. In clinical psychology, somatic symptoms can be divided into typical and atypical forms, and prior research has shown that the common East–West distinction in somatization does not necessarily apply to atypical somatic symptoms \cite{dere2013beyond}. However, because our goal is to assess how well LLMs understand and replicate standardized diagnostic frameworks, we followed the DSM-5, which does not distinguish between these subtypes. This constraint reduced the feasibility of more detailed somatic symptom analyses.

We again acknowledge the limitations of simplifying cultural and symptom categorizations, specifically, the binary distinctions between Western and Eastern, and psychological and somatic. Nevertheless, we view this work as a foundational step, serving as a crucial benchmark for measuring the depth of LLMs' cultural reasoning in relation to depression symptoms. Future research can aim to pursue more nuanced, symptom-level analyses and move beyond these dichotomies.

\section*{Ethical Considerations}
We acknowledge that the binary framework of Western and Eastern cultures captures only a limited portion of the world’s cultural diversity. Many cultures do not fit neatly into this framework. For instance, how individuals from African and Latin American regions express symptoms of depression remains underexplored. Broadly, social psychology has historically emphasized East–West comparisons, often overlooking other cultural contexts \cite{kitayama2024cultural}. We argue that continued collaboration between researchers in computer science and psychology is essential to ensure that LLMs developed for mental health applications are culturally inclusive and effective across diverse populations. 


\section*{Acknowledgments}

This research was supported by the Republic of Korea's MSIT (Ministry of Science and ICT), under the Global Research Support Program in the Digital Field Program (RS-2024-00425354) supervised by the IITP (Institute of Information and Communications Technology Planning \& Evaluation).



\bibliography{custom}

\appendix
\section{Appendix}
\subsection{Additional Details for Experiments}
Experiments with Llama, Qwen, Gemma, and DeepSeek were conducted on NVIDIA A100 GPUs. GPT experiments were conducted using OpenAI’s API.

\subsection{Notation Summary}

\begin{table*}[h]
\centering
\begin{tabular}{ll}
\toprule
\textbf{Notation} & \textbf{Description} \\
\midrule
$x$ & Prompt type: $x \in \{\text{I} \text{ (Implicit)}, \text{E} \text{ (Explicit)}\}$ \\
$c, c_1, c_2$ & Cultural personas (countries) \\
$s$ & Depression symptom (somatic or psychological) \\
$P^x_c$ & Prompt assigned to persona $c$ under prompt type $x \in \{\text{I}, \text{E}\}$ \\
$P(s \mid P^x_c)$ & Probability of selecting symptom $s$ given persona $c$ and prompt type $x$ \\
$P_\Delta(\textsc{p}_x^{(c_1, c_2)}, s)$ & Attribution bias of symptom $s$ between $c_1$ and $c_2$ under prompt $x$ \\
$_1\mathcal{A}_x$ & Cultural alignment score in symptom selection \\
$_2\mathcal{A}_x$ & Cultural alignment vector in cultural attribution \\
$_2\mathcal{A}_x^{\text{som}}$ & Somatic component of $_2\mathcal{A}_x$ \\
$_2\mathcal{A}_x^{\text{psy}}$ & Psychological component of $_2\mathcal{A}_x$ \\
\bottomrule
\end{tabular}
\caption{Notation summary}
\label{tab:notation_summary}
\end{table*}

Table~\ref{tab:notation_summary} summarizes the notations used throughout this paper.

\subsection{Prompt Template}
\label{sec:prompt}
The template of \(\textsc{p}_{I} \) and \(\textsc{p}_{E} \). To ensure LLMs correctly understood each symptom defined in DSM-5, we provided their brief descriptions in the prompts. These descriptions were derived from the PsySym dataset \cite{zhang2022symptom}, where the symptom categories and their associated descriptions were constructed by merging DSM-5 criteria and clinical questionnaire items.
\begin{tcolorbox}[
  colback=gray!10!white, 
  colframe=teal!80!black, 
  sharp corners=southwest, 
  title=\(\textsc{p}_{I} \), 
  coltitle=black, 
  fonttitle=\bfseries,
  listing only,
  listing options={
    basicstyle=\ttfamily,
    breaklines=true
  }
]
You are a (country) person with depression. As a (country) person with depression, select (number of symptoms) symptom you are experiencing from the following symptom list.

\vspace{1em}
(symptom list)
\end{tcolorbox}

\begin{tcolorbox}[
  colback=gray!10!white, 
  colframe=teal!80!black, 
  sharp corners=southwest, 
  title=\(\textsc{p}_{E} \), 
  coltitle=black, 
  fonttitle=\bfseries,
  listing only,
  listing options={
    basicstyle=\ttfamily,
    breaklines=true
  }
]
You are a (country) person with depression. As a (country) person with depression, select (number of symptoms) symptom you are experiencing from the following symptom list. Pay close attention to your cultural background and think carefully about which symptom are likely expressed among the people of your cultural background.

\vspace{1em}
(symptom list)
\end{tcolorbox}

\subsection{The Overall Results}
\label{sec:overall}
\begin{figure*}[h]
  \centering
  \includegraphics[width=0.9\textwidth]{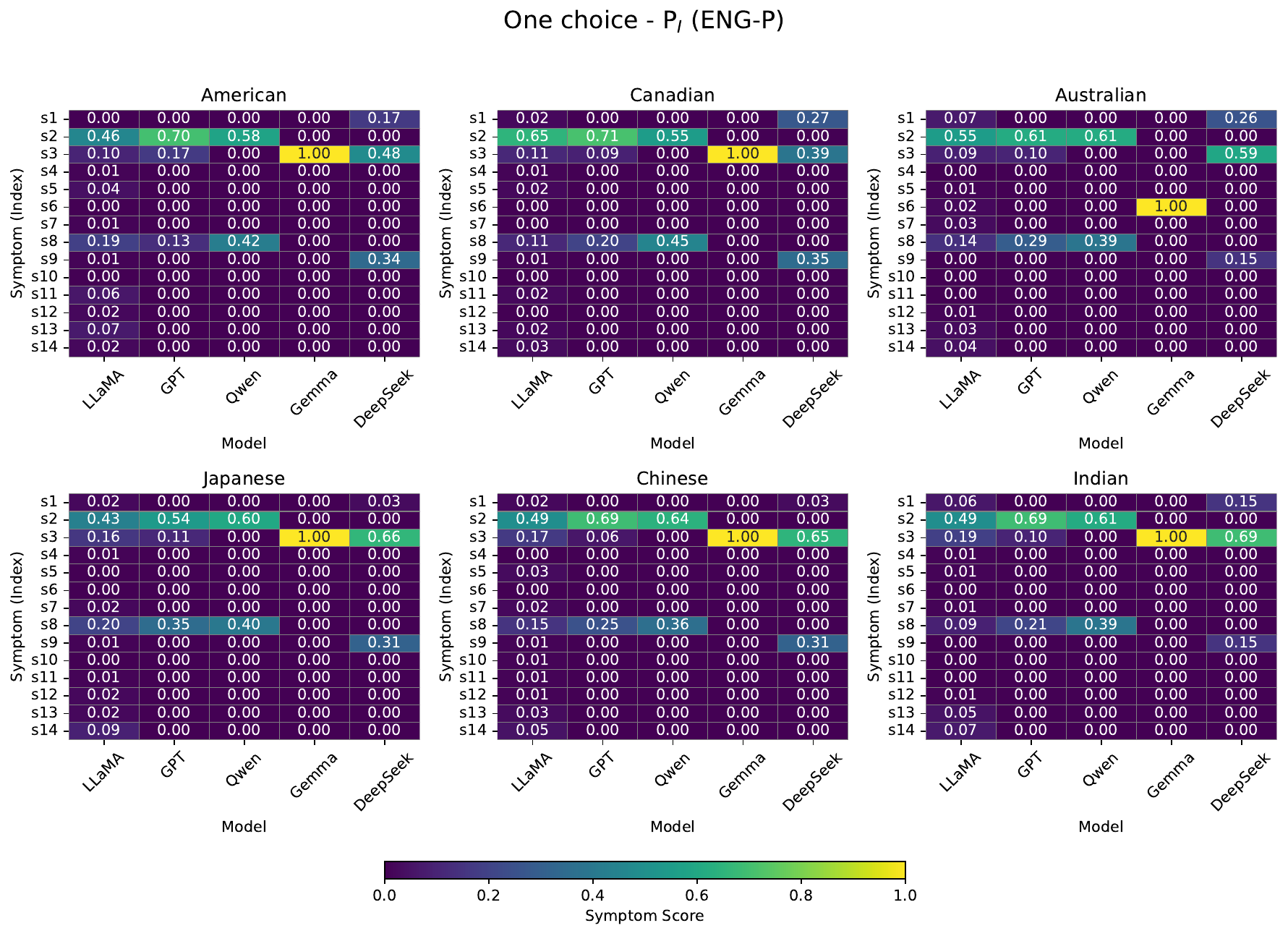}
  \includegraphics[width=0.9\textwidth]{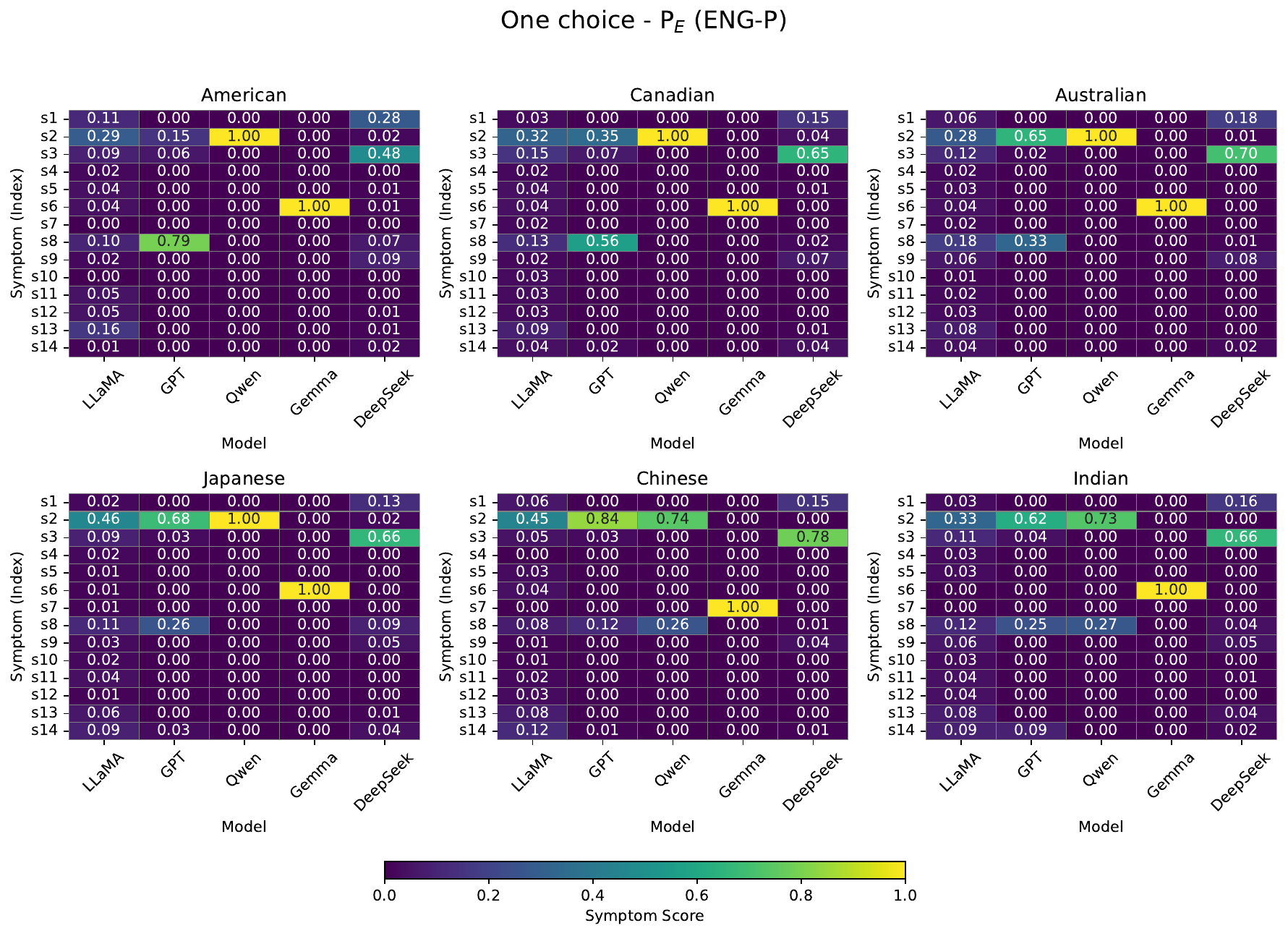}
  \caption{Selected symptom proportions \(P(s \mid \textsc{p}_x^c)\) across models for six cultural personas under one choice condition under the ENG-P condition.}
  \label{fig:combined-heatmaps-one}
\end{figure*}

\begin{figure*}[h]
  \centering
  \includegraphics[width=0.9\textwidth]{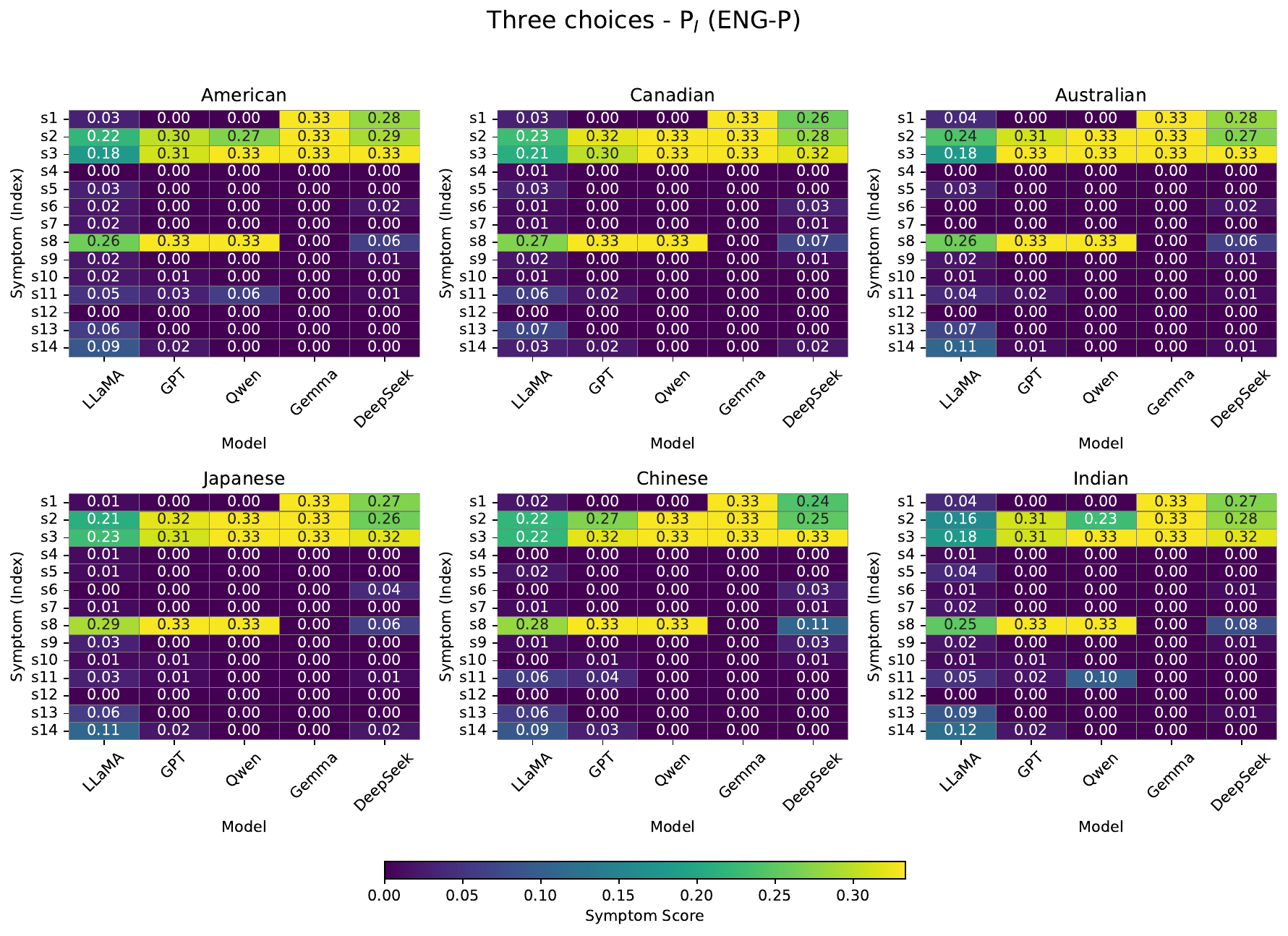}
  \includegraphics[width=0.9\textwidth]{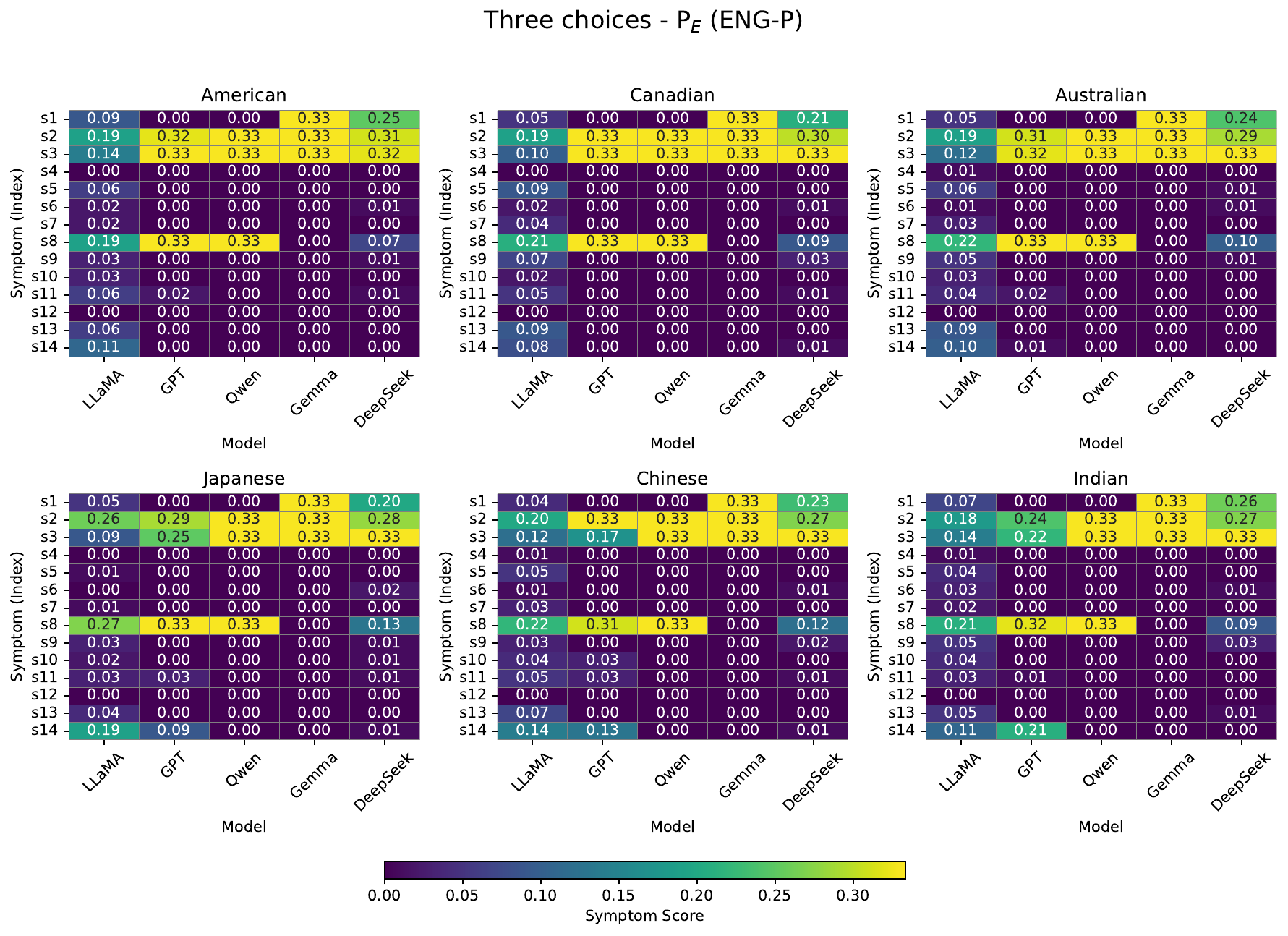}
  \caption{Selected symptom proportions \(P(s \mid \textsc{p}_x^c)\) across models for six cultural personas under three choice condition under the ENG-P condition.}
  \label{fig:combined-heatmaps-three}
\end{figure*}

\begin{figure*}[h]
  \centering
  \includegraphics[width=0.9\textwidth]{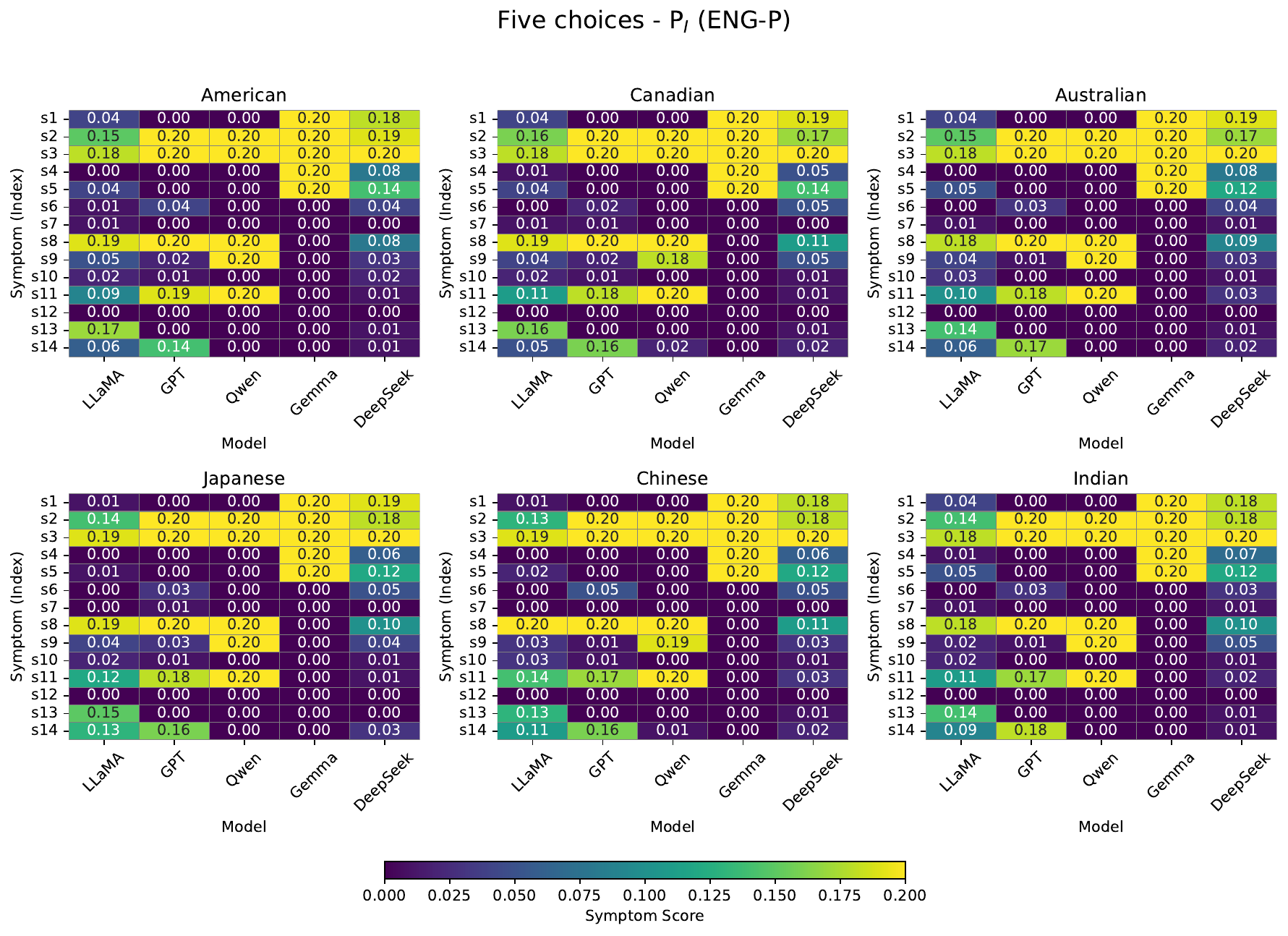}
  \includegraphics[width=0.9\textwidth]{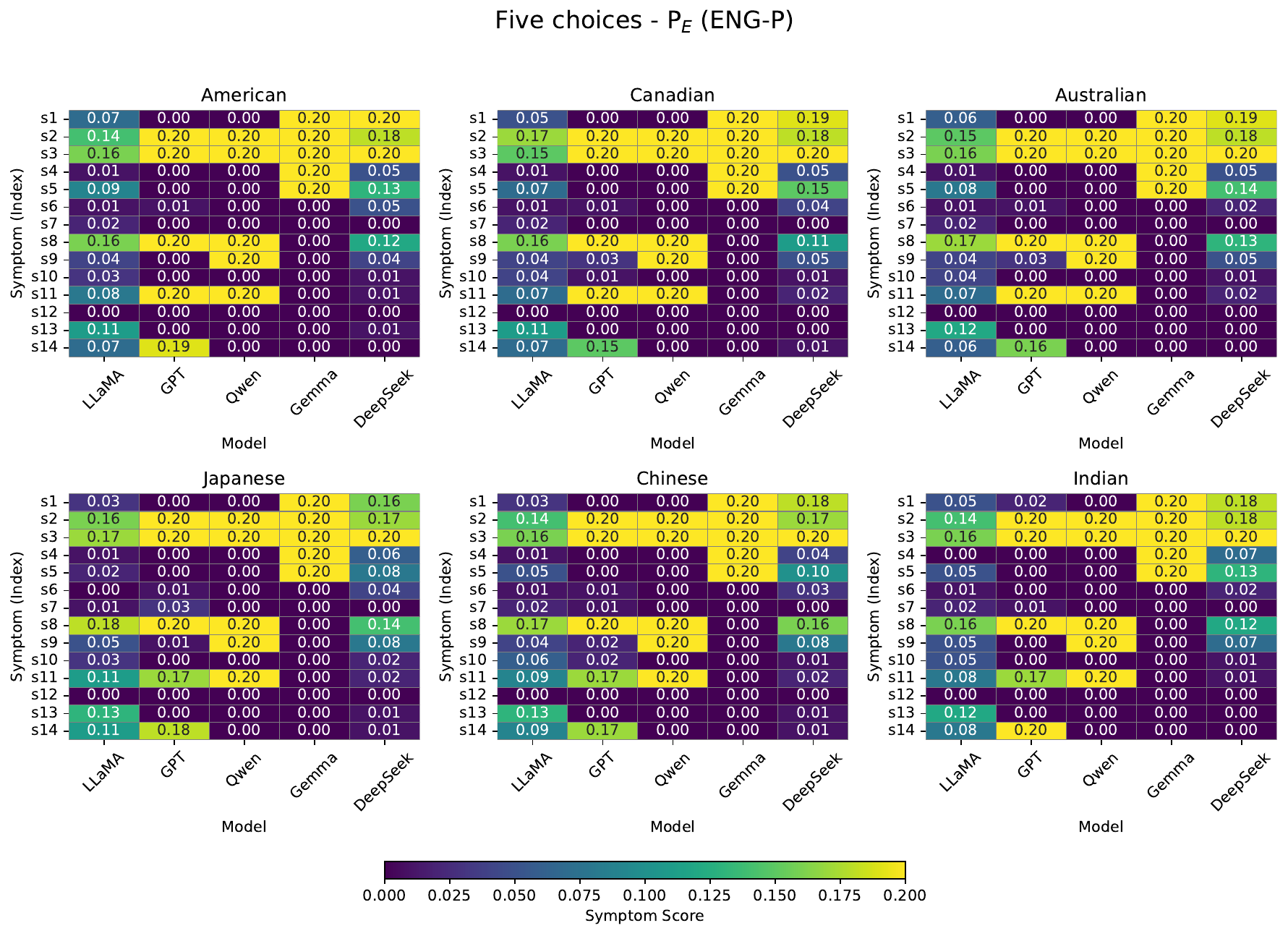}
  \caption{Selected symptom proportions \(P(s \mid \textsc{p}_x^c)\) across models for six cultural personas under five choice condition under the ENG-P condition.}
  \label{fig:combined-heatmaps-five}
\end{figure*}

\begin{figure*}[h]
  \centering
  \includegraphics[width=0.9\textwidth]{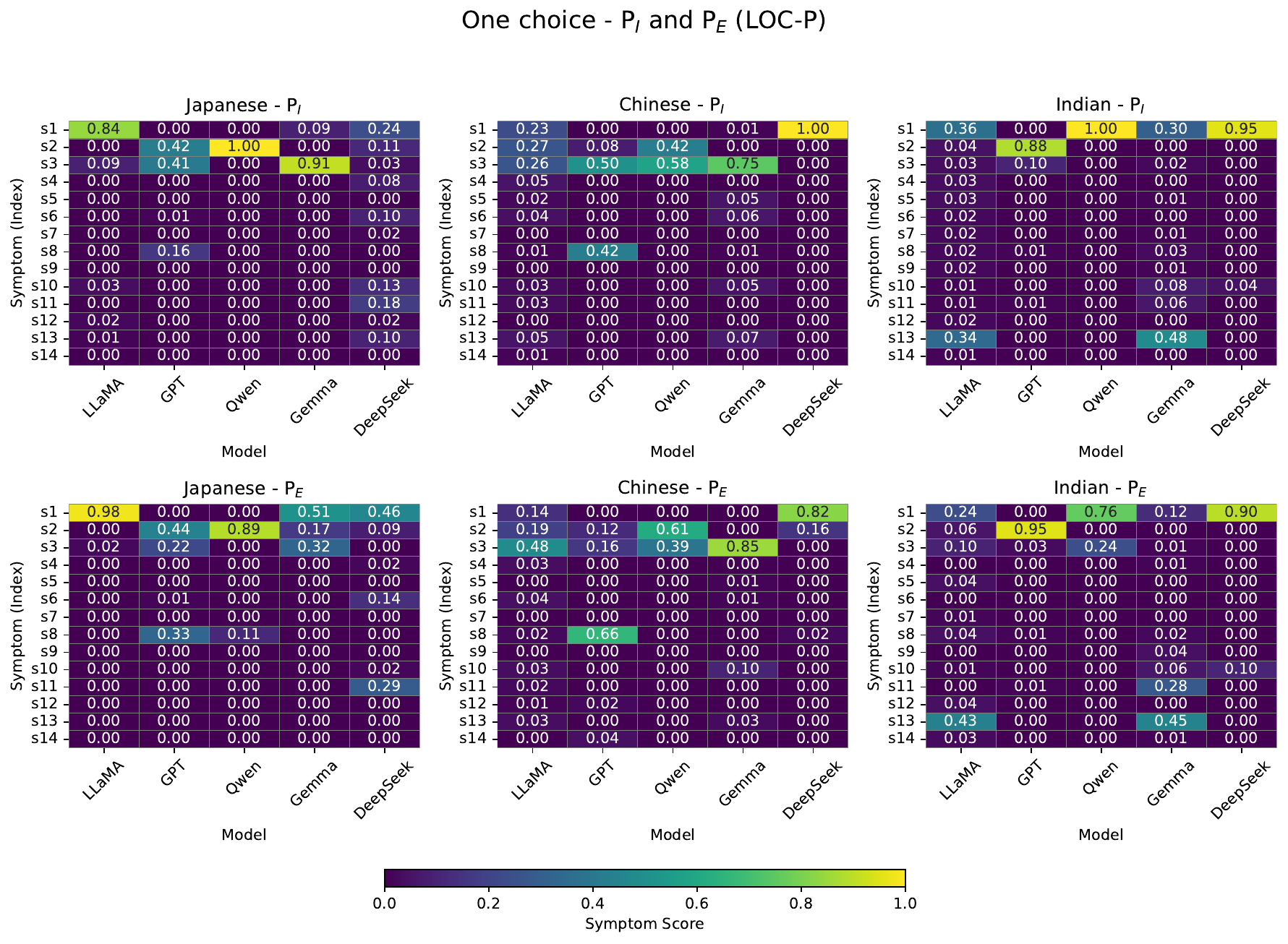}
  \caption{Selected symptom proportions \(P(s \mid \textsc{p}_x^c)\) across models for Eastern cultural personas under one choice condition under the LOC-P condition.}
  \label{fig:combined-heatmaps-one-eastern}
\end{figure*}

\begin{figure*}[h]
  \centering
  \includegraphics[width=0.9\textwidth]{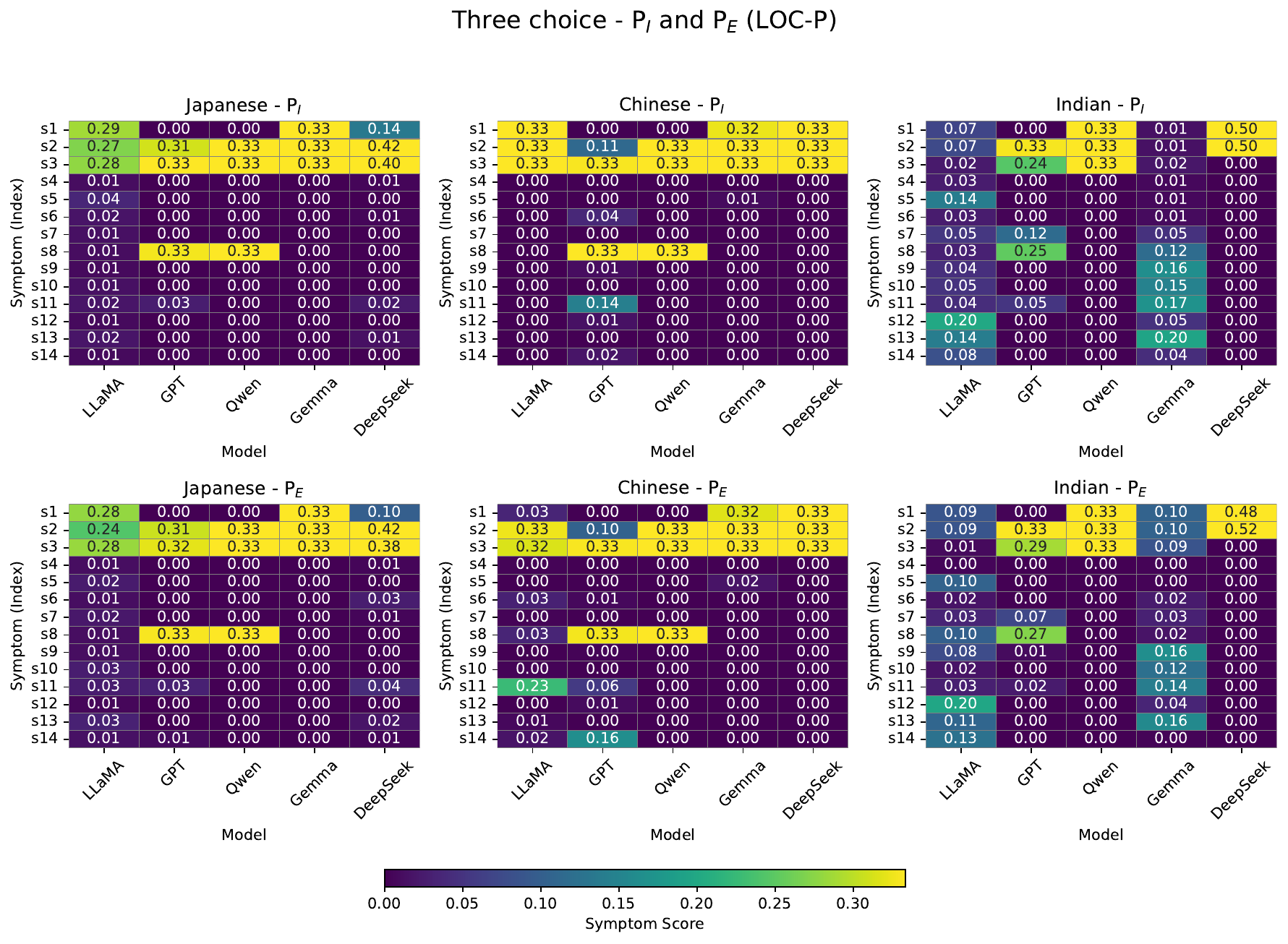}
  \caption{Selected symptom proportions \(P(s \mid \textsc{p}_x^c)\) across models for Eastern cultural personas under three choice condition under the LOC-P condition.}
  \label{fig:combined-heatmaps-three-eastern}
\end{figure*}

\begin{figure*}[h]
  \centering
  \includegraphics[width=0.9\textwidth]{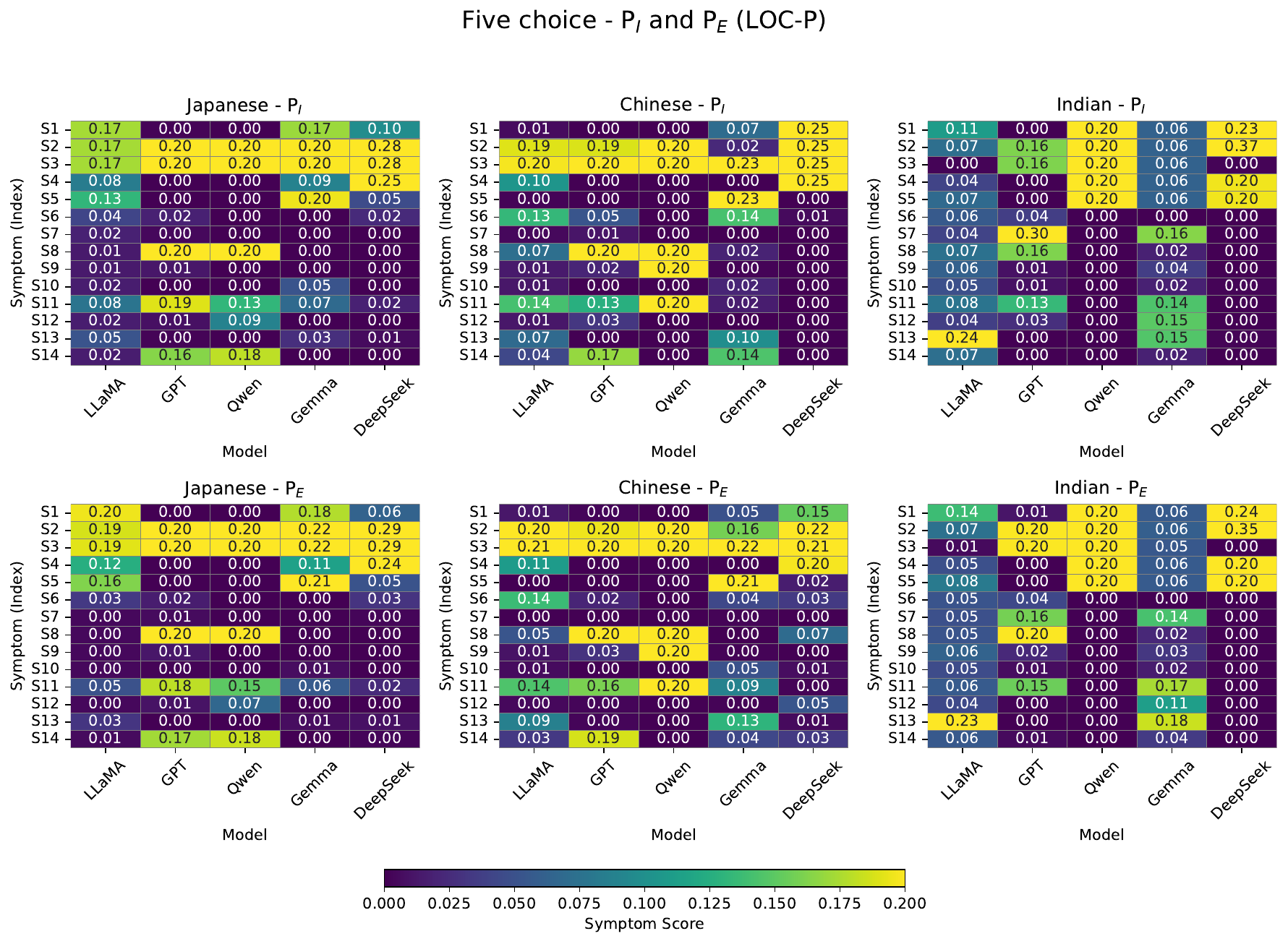}
  \caption{Selected symptom proportions \(P(s \mid \textsc{p}_x^c)\) across models for Eastern cultural personas under five choice condition under the LOC-P condition.}
  \label{fig:combined-heatmaps-five-eastern}
\end{figure*}

Figures \ref{fig:combined-heatmaps-one} to \ref{fig:combined-heatmaps-five-eastern} display the proportions of selected symptoms \(P(s \mid \textsc{p}_x^c)\) across all the experimental settings under the ENG-P condition (Figures \ref{fig:combined-heatmaps-one}, \ref{fig:combined-heatmaps-three}, and \ref{fig:combined-heatmaps-five}) and under the LOC-P condition (Figures \ref{fig:combined-heatmaps-one-eastern}, \ref{fig:combined-heatmaps-three-eastern}, and \ref{fig:combined-heatmaps-five-eastern}). 

\subsection{Alignment}
\label{sec:alignment}
Table \ref{tab:symptom-region-model-all-en-task1} (under the ENG-P condition) and \ref{tab:symptom-region-model-all-dif} (under the LOC-P condition) present the proportions of somatic symptoms \( S_{\text{som}} \) and psychological symptoms \( S_{\text{psy}} \) selected by Western and Eastern personas across all experimental settings. It is averaged within somatic or psychological symptoms and also within Western or Eastern personas. Table \ref{tab:x_diff_table} presents $_1\mathcal{A}_x(l(c)) - {_1}\mathcal{A}_x(Eng)$ values across models by experimental settings. 

\begin{table*}[h]
\centering
\resizebox{\textwidth}{!}{%
\begin{tabular}{llcccccccccc}
\toprule
\textbf{Condition} & \textbf{Symptom Type}
& \multicolumn{2}{c}{\textbf{Llama}}
& \multicolumn{2}{c}{\textbf{GPT}}
& \multicolumn{2}{c}{\textbf{Qwen}}
& \multicolumn{2}{c}{\textbf{Gemma}}
& \multicolumn{2}{c}{\textbf{DeepSeek}} \\
&
& Western & Eastern
& Western & Eastern
& Western & Eastern
& Western & Eastern
& Western & Eastern \\
\midrule
One choice \(\textsc{p}_{I} \) & Somatic       & 0.65(0.24) & 0.54(0.20) & 0.67(0.30) & 0.64(0.29) & \textbf{0.58(0.26)} & \textbf{0.62(0.28)} & 0.00(0.00) & 0.00(0.00) & 0.00(0.00) & 0.00(0.00) \\ 
One choice \(\textsc{p}_{I} \) & Psychological & 0.34(0.05) & 0.46(0.07) & 0.33(0.08) & 0.36(0.09) & \textbf{0.42(0.14)} & \textbf{0.38(0.13)} & 1.00(0.24) & 1.00(0.33) & 1.00(0.18) & 1.00(0.23) \\ 
One choice \(\textsc{p}_{E} \) & Somatic       & \textbf{0.49(0.11)} & \textbf{0.57(0.17)} & \textbf{0.38(0.17)} & \textbf{0.73(0.32)} & 1.00(0.45) & 0.82(0.13) & 0.00(0.00) & 0.00(0.00) & 0.05(0.01) & 0.03(0.01) \\ 
One choice \(\textsc{p}_{E} \) & Psychological & \textbf{0.49(0.05)} & \textbf{0.43(0.04)} & \textbf{0.62(0.18)} & \textbf{0.28(0.07)} & 0.00(0.00) & 0.18(0.06) & 1.00(0.33) & 1.00(0.24) & 0.95(0.20) & 0.97(0.23) \\ 
Three choice \(\textsc{p}_{I} \) & Somatic       & 0.38(0.09) & 0.35(0.08) & 0.34(0.14) & 0.33(0.13) & 0.33(0.14) & 0.33(0.13) & 0.33(0.15) & 0.33(0.15) & 0.29(0.12) & 0.29(0.12) \\ 
Three choice \(\textsc{p}_{I} \) & Psychological & 0.61(0.09) & 0.65(0.10) & 0.66(0.14) & 0.67(0.14) & 0.67(0.15) & 0.67(0.15) & 0.67(0.15) & 0.67(0.15) & 0.71(0.13) & 0.71(0.12) \\ 
Three choice \(\textsc{p}_{E} \) & Somatic       & 0.40(0.07) & 0.35(0.08) & 0.34(0.14) & 0.32(0.13) & 0.33(0.15) & 0.33(0.15) & 0.33(0.15) & 0.33(0.15) & 0.32(0.13) & 0.29(0.12) \\ 
Three choice \(\textsc{p}_{E} \) & Psychological & 0.61(0.07) & 0.65(0.08) & 0.66(0.15) & 0.68(0.12) & 0.67(0.15) & 0.67(0.15) & 0.67(0.15) & 0.67(0.15) & 0.68(0.12) & 0.71(0.12) \\ 
Five choice \(\textsc{p}_{I} \) & Somatic       & 0.45(0.07) & 0.42(0.06) & 0.38(0.10) & 0.37(0.10) & 0.40(0.11) & 0.40(0.11) & 0.60(0.11) & 0.60(0.11) & 0.40(0.07) & 0.39(0.07) \\ 
Five choice \(\textsc{p}_{I} \) & Psychological & 0.54(0.07) & 0.58(0.08) & 0.62(0.09) & 0.63(0.09) & 0.60(0.10) & 0.60(0.10) & 0.40(0.09) & 0.40(0.09) & 0.60(0.08) & 0.61(0.08) \\ 
Five choice \(\textsc{p}_{E} \) & Somatic       & 0.44(0.05) & 0.40(0.06) & 0.40(0.11) & 0.37(0.10) & 0.40(0.11) & 0.40(0.11) & 0.60(0.11) & 0.60(0.11) & 0.40(0.08) & 0.37(0.07) \\ 
Five choice \(\textsc{p}_{E} \) & Psychological & 0.56(0.06) & 0.60(0.07) & 0.60(0.09) & 0.63(0.09) & 0.60(0.10) & 0.60(0.10) & 0.40(0.09) & 0.40(0.09) & 0.60(0.08) & 0.63(0.08) \\
\bottomrule
\end{tabular}
}
\caption{Average proportions of selected symptoms by type (\( S_{\text{som}} \) vs. \( S_{\text{psy}} \)) and cultural personas (\(\mathcal{C}_W\) vs. \(\mathcal{C}_E\)) for each experimental setting under the ENG-P condition. Values
in parentheses indicate standard deviations. Experimental settings demonstrating cultural alignment are highlighted in bold.}
\label{tab:symptom-region-model-all-en-task1}
\end{table*}

\begin{table*}[h]
\centering
\resizebox{\textwidth}{!}{%
\begin{tabular}{llcccccccccc}
\toprule
\textbf{Condition} & \textbf{Symptom Type}
& \multicolumn{2}{c}{\textbf{Llama}}
& \multicolumn{2}{c}{\textbf{GPT}}
& \multicolumn{2}{c}{\textbf{Qwen}}
& \multicolumn{2}{c}{\textbf{Gemma}}
& \multicolumn{2}{c}{\textbf{DeepSeek}} \\
&
& Western & Eastern
& Western & Eastern
& Western & Eastern
& Western & Eastern
& Western & Eastern \\
\midrule
One choice \(\textsc{p}_{I} \) & Somatic         & 0.65(0.24) & 0.28(0.06) & 0.67(0.30) & 0.46(0.21) & 0.58(0.26) & 0.47(0.21) & \textbf{0.00(0.00)} & \textbf{0.23(0.08)} & \textbf{0.00(0.00)} & \textbf{0.15(0.02)} \\
One choice \(\textsc{p}_{I} \) & Psychological   & 0.34(0.05) & 0.72(0.15) & 0.33(0.08) & 0.54(0.12) & 0.42(0.14) & 0.53(0.12) & \textbf{1.00(0.24)} & \textbf{0.77(0.18)} & \textbf{1.00(0.18)} & \textbf{0.85(0.24)} \\
One choice \(\textsc{p}_{E} \) & Somatic         & 0.49(0.11) & 0.00(0.00) & \textbf{0.38(0.17)} & \textbf{0.51(0.22)} & 1.00(0.45) & 0.50(0.22) & \textbf{0.00(0.00)} & \textbf{0.36(0.07)} & \textbf{0.05(0.01)} & \textbf{0.17(0.05)} \\
One choice \(\textsc{p}_{E} \) & Psychological   & 0.49(0.05) & 1.00(0.15) & \textbf{0.62(0.18)} & \textbf{0.49(0.11)} & 0.00(0.00) & 0.50(0.10) & \textbf{1.00(0.33)} & \textbf{0.64(0.14)} & \textbf{0.95(0.20)} & \textbf{0.83(0.24)} \\
Three choice \(\textsc{p}_{I} \) & Somatic       & 0.38(0.09) & 0.38(0.09) & 0.34(0.14) & 0.32(0.11) & 0.33(0.14) & 0.33(0.15) & \textbf{0.33(0.15)} & \textbf{0.48(0.09)} & \textbf{0.29(0.12)} & \textbf{0.44(0.19)} \\
Three choice \(\textsc{p}_{I} \) & Psychological & 0.62(0.09) & 0.62(0.09) & 0.66(0.14) & 0.68(0.13) & 0.67(0.15) & 0.67(0.12) & \textbf{0.67(0.15)} & \textbf{0.52(0.09)} & \textbf{0.71(0.13)} & \textbf{0.56(0.13)} \\
Three choice \(\textsc{p}_{E} \) & Somatic       & \textbf{0.40(0.07)} & \textbf{0.41(0.08)} & 0.33(0.14) & 0.28(0.11) & 0.33(0.15) & 0.33(0.15) & 0.33(0.15) & 0.32(0.11) & \textbf{0.32(0.13)} & \textbf{0.44(0.19)} \\
Three choice \(\textsc{p}_{E} \) & Psychological & \textbf{0.60(0.07)} & \textbf{0.59(0.06)} & 0.67(0.15) & 0.72(0.13) & 0.67(0.15) & 0.67(0.12) & 0.67(0.15) & 0.68(0.10) & \textbf{0.68(0.12)} & \textbf{0.56(0.12)} \\
Five choice \(\textsc{p}_{I} \) & Somatic        & \textbf{0.45(0.07)} & \textbf{0.51(0.03)} & 0.38(0.10) & 0.33(0.09) & \textbf{0.40(0.11)} & \textbf{0.45(0.07)} & 0.60(0.11) & 0.37(0.04) & \textbf{0.40(0.07)} & \textbf{0.62(0.13)} \\
Five choice \(\textsc{p}_{I} \) & Psychological  & \textbf{0.55(0.07)} & \textbf{0.49(0.04)} & 0.62(0.09) & 0.67(0.08) & \textbf{0.60(0.10)} & \textbf{0.55(0.07)} & 0.40(0.09) & 0.63(0.05) & \textbf{0.60(0.08)} & \textbf{0.38(0.08)} \\
Five choice \(\textsc{p}_{E} \) & Somatic        & \textbf{0.44(0.05)} & \textbf{0.52(0.03)} & 0.40(0.11) & 0.36(0.10) & \textbf{0.40(0.11)} & \textbf{0.45(0.07)} & 0.60(0.11) & 0.58(0.04) & \textbf{0.40(0.08)} & \textbf{0.60(0.13)} \\
Five choice \(\textsc{p}_{E} \) & Psychological  & \textbf{0.56(0.06)} & \textbf{0.48(0.04)} & 0.60(0.09) & 0.64(0.08) & \textbf{0.60(0.10)} & \textbf{0.55(0.07)} & 0.40(0.09) & 0.42(0.05) & \textbf{0.60(0.08)} & \textbf{0.39(0.07)} \\
\bottomrule
\end{tabular}
}
\caption{Average proportions of selected symptoms by type (\( S_{\text{som}} \) vs. \( S_{\text{psy}} \)) and cultural personas (\(\mathcal{C}_W\) vs. \(\mathcal{C}_E\)) for each experimental setting under LOC-P condition. Values
in parentheses indicate standard deviations. Experimental settings demonstrating cultural alignment are highlighted in bold.}
\label{tab:symptom-region-model-all-dif}
\end{table*}

\begin{table}[t]
\centering
\resizebox{\columnwidth}{!}{%
\begin{tabular}{llccccc}
\toprule
\textbf{Choice Condition} & \textbf{Prompt} & \textbf{Llama} & \textbf{GPT} & \textbf{Qwen} & \textbf{Gemma} & \textbf{DeepSeek} \\
\midrule
One choice   & I & -0.26 & -0.18 & -0.15 &  \textbf{0.23} &  \textbf{0.15} \\
One choice   & E & -0.57 & -0.22 & -0.32 &  \textbf{0.36} &  \textbf{0.14} \\
Three choice & I &  \textbf{0.03} & -0.01 &  0.00 &  \textbf{0.15} &  \textbf{0.15} \\
Three choice & E &  \textbf{0.06} & -0.03 &  0.00 & -0.01 &  \textbf{0.15} \\
Five choice  & I &  \textbf{0.09} & -0.04 &  \textbf{0.05} & -0.23 &  \textbf{0.23} \\
Five choice  & E &  \textbf{0.12} & -0.01 &  \textbf{0.05}& -0.02 &  \textbf{0.23} \\
\bottomrule
\end{tabular}
}
\caption{$_1\mathcal{A}_x(l(c)) - {_1}\mathcal{A}_x(Eng)$ values across models by experimental settings. Bolded settings indicate improved alignment. }
\label{tab:x_diff_table}
\end{table}

\subsection{Determinism in Symptom Selection}
\label{sec:determinismsymptom}
To statistically assess each model's degree of determinism in symptom selection, we calculate the average Gini coefficient of \(P(s\mid\textsc{p}_x^c)\) distributions across countries for each experimental setting under the ENG-P and LOC-P condition (Table \ref{tab:gini} and \ref{tab:gini_east}). Higher Gini coefficients indicate the model concentrates selections on a few specific symptoms, while lower values indicate more diverse responses. This analysis has important implications for real-world applications: overly deterministic models may overlook less common but clinically relevant symptoms. 

We observe that different LLMs exhibit distinct behaviors in symptom selection. Across experimental settings, Llama consistently exhibits the lowest Gini coefficients (0.56 on average), indicating greater diversity in its outputs. In contrast, Gemma and Qwen tend to show the highest Gini coefficients (0.78 and 0.77 on average), repeatedly selecting a limited set of symptoms. This suggests stronger inherent preferences for certain depression symptoms. Consistent with the ENG-P condition, Llama consistently shows the lowest Gini coefficient values, showing more diverse symptom selection behaviors. However, unlike ENG-P condition, the DeepSeek consistently exhibit the most deterministic behavior with five out of six experiments. 

The narrow symptom range observed in Gemma, Qwen, and DeepSeek raises concerns, as it may lead to underdiagnosis by missing culturally specific symptom expressions. 
This is especially problematic in multicultural contexts, where depression may manifest differently. While determinism can be useful when aligned with cultural norms, the poor alignment shown by Gemma and Qwen suggests that their determinism is more harmful than helpful.
In contrast, Llama's higher variability may be advantageous for applications requiring flexible and culturally responsive AI. However, excessive variability can also increase the risk of generating irrelevant or inconsistent outputs. Therefore, determining the appropriate level of variability may require input from domain experts. While the exact reasons for these model-specific determinism levels warrant further investigation, they might relate to differences in pre-training data diversity, model architecture, or fine-tuning objectives. 

\setcounter{table}{8}
\begin{table}[h]
    \centering
    \resizebox{\columnwidth}{!}{%
    \begin{tabular}{lrrrrr}
        \toprule
        & \textbf{Llama} & \textbf{GPT} & \textbf{Qwen} & \textbf{Gemma} & \textbf{DeepSeek} \\
        \midrule
        One choice \(\textsc{p}_{I} \) & \underline{0.72}(0.02) & 0.87(0.01) & 0.87(0.00) & \textbf{0.92}(0.02) & 0.85(0.03) \\
        One choice \(\textsc{p}_{E} \) & \underline{0.54}(0.05) & 0.87(0.02) & \textbf{0.92}(0.02) & \textbf{0.92}(0.02) & 0.83(0.03) \\
        Three choice \(\textsc{p}_{I} \) & \underline{0.61}(0.03) & 0.77(0.00) & 0.78(0.00) & \textbf{0.79}(0.00) & 0.74(0.01) \\
        Three choice \(\textsc{p}_{E} \) & \underline{0.50}(0.05) & 0.76(0.02) & \textbf{0.79}(0.00) & \textbf{0.79}(0.00) & 0.74(0.01) \\
        Five choice \(\textsc{p}_{I} \) & \underline{0.53}(0.02) & 0.63(0.01) & \textbf{0.64}(0.00) & \textbf{0.64}(0.00) & 0.54(0.01) \\
        Five choice \(\textsc{p}_{E} \) & \underline{0.45}(0.02) & 0.63(0.01) & \textbf{0.64}(0.00) & \textbf{0.64}(0.00) & 0.55(0.02) \\
        Average  & \underline{0.56} & 0.76 & 0.77 & \textbf{0.78} & 0.71 \\
        \bottomrule
    \end{tabular}
    }
    \caption{The average Gini coefficient for each model under the ENG-P condition. The highest values are highlighted in bold, while the lowest are underlined. Values in parentheses indicate standard deviations across six countries.}
    \label{tab:gini}
\end{table}

\begin{table}[h]
    \centering
    \resizebox{\columnwidth}{!}{%
    \begin{tabular}{lrrrrr}
        \toprule
        & \textbf{Llama} & \textbf{GPT} & \textbf{Qwen} & \textbf{Gemma} & \textbf{DeepSeek} \\
        \midrule
        One choice \(\textsc{p}_{I} \) & \underline{0.70}(0.11) & 0.84(0.04) & \textbf{0.88}(0.05) & 0.82(0.06) & 0.80(0.15) \\
        One choice \(\textsc{p}_{E} \) & \underline{0.75}(0.10) & 0.85(0.04) & \textbf{0.86}(0.04) & 0.78(0.08) & \textbf{0.86}(0.05) \\
        Three choice \(\textsc{p}_{I} \) & \underline{0.58}(0.15) & 0.74(0.02) & 0.78(0.01) & 0.65(0.13) & \textbf{0.81}(0.03) \\
        Three choice \(\textsc{p}_{E} \) & \underline{0.54}(0.13) & 0.75(0.02) & 0.78(0.01) & 0.66(0.13) & \textbf{0.80}(0.03) \\
        Five choice \(\textsc{p}_{I} \) & \underline{0.39}(0.10) & 0.60(0.03) & 0.58(0.08) & 0.46(0.13) & \textbf{0.72}(0.02) \\
        Five choice \(\textsc{p}_{E} \) & \underline{0.44}(0.12) & 0.61(0.01) & 0.58(0.07) & 0.50(0.10) & \textbf{0.67}(0.06) \\
        \bottomrule
    \end{tabular}
    }
    \caption{The average Gini coefficient for each model under the LOC-P condition. The highest values are highlighted in bold, while the lowest are underlined. Values in parentheses indicate standard deviations across six countries.}
    \label{tab:gini_east}
\end{table}

\subsection{Symptom Selection Differences between Western and Eastern Personas}
\label{sec:symptomselectiondifferences}

Table \ref{tab:statistically_significant_symptoms} and \ref{tab:statistically_significant_symptoms_east} list the symptoms that show statistically significant differences (p < 0.05) between Western and Eastern personas under the ENG-P and LOC-P. We performed chi-square tests between \(P(s \mid \textsc{p}_x^\text{Western})\) and \(P(s \mid \textsc{p}_x^\text{Eastern})\) under each experimental condition. Under the ENG-P condition, no single symptom consistently emerges as significant across the models, suggesting that cultural bias in symptom selection is not robust. Furthermore, the set of statistically significant symptoms varies by prompt type and choice condition, even \emph{within} the same model. One notable exception is s14 (Worthlessness and guilt; Psychological), which Llama consistently selects more frequently for Eastern personas across all conditions. This pattern suggests that Llama may have learned a strong association between Eastern cultural personas and the symptom of worthlessness and guilt. Interestingly, this diverges from prior clinical psychology findings, which typically report a greater emphasis of somatic symptoms in Eastern populations. GPT exhibits significant differences only under \(\textsc{p}_{E} \), suggesting that it may require more explicit instructions to reflect Western-Eastern distinctions in its outputs. Under the LOC-P condition, we observe a greater number of symptoms with statistically significant differences between Western and Eastern personas, demonstrating the effectiveness of prompt language to reflect Western-Eastern distinctions. However, as it is shown in \S\ref{sec:languageeffect}, these distinctions do not reflect clinically expected cultural patterns.

\begin{table*}[h]
\centering
\resizebox{\textwidth}{!}{%
\begin{tabular}{lccccccccccc}
\toprule
\textbf{Setting} & \multicolumn{2}{c}{Llama} & \multicolumn{2}{c}{GPT} & \multicolumn{2}{c}{Qwen} & \multicolumn{2}{c}{Gemma} & \multicolumn{2}{c}{DeepSeek} \\
& Western & Eastern & Western & Eastern & Western & Eastern & Western & Eastern & Western & Eastern \\
\midrule
One choice \(\textsc{p}_{I} \) & & s3,s14 &  &  &  &  & s6 & s3 & s1 & s3 \\
One choice \(\textsc{p}_{E} \) & & s2,s14 & s8 & s2,s14 & s3 & s8 & s6 & s7 &  & s3 \\ 
Three choice \(\textsc{p}_{I} \) & & s14 &  &  &  &  &  &  &  &  \\
Three choice \(\textsc{p}_{E} \) &s5,s13 & s14 & s3 & s10,s14 &  &  &  &  &  &  \\
Five choice \(\textsc{p}_{I} \) &s1,s5,s9 & s14 &  &  &  &  &  &  &  &  \\
Five choice \(\textsc{p}_{E} \) &s1,s5 & s14 & s11 & s1,s7 &  &  &  &  & s5 & s9 \\
\bottomrule
\end{tabular}
}
\caption{Depression symptoms which showed statistically significant difference between Western and Eastern countries and their directions under the ENG-P condition.}
\label{tab:statistically_significant_symptoms}
\end{table*}

\begin{table*}[h]
\centering
\resizebox{\textwidth}{!}{%
\begin{tabular}{lccccccccccc}
\toprule
\textbf{Setting} & \multicolumn{2}{c}{Llama} & \multicolumn{2}{c}{GPT} & \multicolumn{2}{c}{Qwen} & \multicolumn{2}{c}{Gemma} & \multicolumn{2}{c}{DeepSeek} \\
& Western & Eastern & Western & Eastern & Western & Eastern & Western & Eastern & Western & Eastern \\
\midrule
One choice \(\textsc{p}_{I} \) & s2,s8 & s1,s10,s13 & s2 & s3 & s2,s8 & s1,s3 & s3,s6 & s1,s5,s10,s11,s13 & s3,s9 & s1,s2,s6,s10,s11,s13 \\
One choice \(\textsc{p}_{E} \) & s2,s8,s9 & s1,s3 & s8 & s2,s3 & s2 & s1,s3,s8 & s6 & s1,s2,s3,s10,s11,s13 & s3,s9,s14 & s1,s2,s6,s10,s11 \\ 
Three choice \(\textsc{p}_{I} \) & s8,s14 & s1,s4,s5,s7,s12 & s2 & s6,s7,s11 & s8,s11 & s1 & s1–s3 & s7–s14 & s3,s6,s8,s9,s14 & s2 \\
Three choice \(\textsc{p}_{E} \) & s5,s8,s9,s13,s14 & s1,s3,s11,s12 & s2 & s7,s11,s14 & s8 & s1 & s1–s3 & s5–s7,s9–s13 & s3,s8,s9 & s1,s2 \\
Five choice \(\textsc{p}_{I} \) & s3,s8,s9,s13,s14 & s1,s4,s5,s6,s7,s12 &  & s7,s12 & s8,s9,s11 & s1,s4,s5,s12,s14 & s1–s5 & s6–s14 & s3,s5,s6,s8–s11,s14 & s2,s4 \\
Five choice \(\textsc{p}_{E} \) & s8–s10 & s1,s4,s6,s12,s14 & s11,s14 & s6,s7,s12 & s8,s9,s11 & s1,s4,s5,s12,s14 & s1,s2,s4,s6 & s7–s14 & s1,s3,s5,s8–s11 & s2,s4,s12,s14 \\
\bottomrule
\end{tabular}
}
\caption{Depression symptoms which showed statistically significant difference between Western and Eastern countries and their directions under the LOC-P condition. }
\label{tab:statistically_significant_symptoms_east}
\end{table*}

\subsection{Example Outputs by MentaLLaMA}
\label{sec:mental-llama}

Table~\ref{tab:mental-llama-examples} shows two example outputs by MentaLLaMA.

\begin{table}[h!]
\centering
\resizebox{\columnwidth}{!}{%
\begin{tabular}{|c|p{10cm}|}
\hline
\textbf{Example 1} &
\textit{Note: The provided symptom list is not exhaustive and may not cover all possible symptoms of depression. It is important to consult a mental health professional for an accurate diagnosis and treatment plan.}
\\ \hline

\textbf{Example 2} &
\textit{Note: The given list of symptoms is not exhaustive and is based on the DSM-5 criteria for depression. It is important to consult with a mental health professional for an accurate diagnosis and treatment plan.}
\\ \hline
\end{tabular}%
}
\caption{Example Outputs by MentaLLaMA}
\label{tab:mental-llama-examples}
\end{table}

\subsection{Individual Symptom Level Analysis}
\label{sec:individual}
We also conducted a more detailed, symptom-level analysis of psychological symptoms. While prior research generally suggests Eastern somatization and Western psycholization, not all psychological symptoms have been consistently reported as more prevalent among Western populations or showed statistically significant differences between Western and Eastern samples \cite{parkerChineseSomatizeDepression2001, biswasCrossCulturalVariations2016, dere2013beyond}. 
We first identified psychological symptoms with statistically significant differences between Western and Eastern populations (p $<$ 0.05). These were drawn from prior studies that conducted symptom-level analyses \cite{parkerChineseSomatizeDepression2001, biswasCrossCulturalVariations2016, dere2013beyond}.
We then matched these symptoms with their corresponding entries in the DSM-5. Symptoms not included in the DSM-5 were excluded from this analysis. Finally, we examined whether LLMs exhibited similar patterns of statistically significant differences (p $<$ 0.05) in symptom selection across Eastern and Western cultural personas. Please refer to \S\ref{sec:limitation} for a discussion of why we did not perform individual-level analyses for somatic symptoms.

\textbf{Australian-Chinese Pair.} A study by \citet{parkerChineseSomatizeDepression2001} found that \emph{Depressed mood} and \emph{Loss of interest} were significantly more frequently reported by Australian patients, whereas \emph{Suicidal thoughts} were more commonly reported by Chinese patients. In our results, alignment with \citet{parkerChineseSomatizeDepression2001} is limited: for \emph{Depressed mood}, only one condition (GPT-LOC-P, One choice) shows alignment. For \emph{Loss of interest}, only the (GPT, LOC-P, Three choice) condition aligns with the clinical findings. \emph{Suicidal thoughts} has no alignment condition. 

\textbf{Canadian-Chinese Pair.} A study by \citet{dere2013beyond} reported that \emph{Depressed mood} was significantly more likely to be expressed by Chinese patients than Canadian patients. We find alignment with this result in three conditions: (Deepseek, ENG-P, One choice), (Llama, LOC-P, One choice), and (GPT, LOC-P, Three choice).

\textbf{American-Indian Pair.} A study by \citet{biswasCrossCulturalVariations2016} found that American psychiatrists placed greater emphasis on \emph{Decreased interest in pleasurable activities}, \emph{Pessimistic view of the future}, and \emph{Ideas of self-harm or suicide}. In our results, \emph{Decreased interest in pleasurable activities} aligns in only one condition: (GPT, LOC-P, One choice). For \emph{Pessimistic view of the future}, two conditions show alignment: (Deepseek, ENG-P, One choice) and (Llama, ENG-P, Five choice). \emph{Ideas of self-harm or suicide} has no alignment condition. 

Overall, across 30 experimental settings (3 choice conditions $\times$ 2 prompt types $\times$ 5 models) for each symptom, only zero to three conditions per symptom showed alignment with prior clinical psychology studies.
Alignment at the individual symptom level for psychological symptoms is limited mainly due to the small absolute values of \(P(s \mid \textsc{p}_x^{\text{Eastern}}) - P(s \mid \textsc{p}_x^{\text{Western}})\).

\section{Appendix for Cultural Attribution Task}
\label{sec:culturalattribution}
\subsection{Research Hypotheses}
We examine whether LLMs show cultural attribution when inferring a cultural group based on given symptoms.

\begin{itemize}[label={}]
    \item \textbf{H1.} LLMs are more likely to attribute psychological symptoms to Western cultural groups and somatic symptoms to Eastern cultural groups.
    \item \textbf{H2.} Prompts written in the local language of a country increase cultural alignment in cultural attribution.
\end{itemize}

\subsection{Task Design for Hypothesis Testing}
We invert the task structure of our main analysis: instead of providing a cultural persona, we present a single symptom (e.g., ``Anger and irritability'') and ask the model to decide whether it is more likely to appear in a person from one of two cultural groups, such as ``American'' or ``Japanese.''

Each prompt compares a Western culture \( c_1 \in \mathcal{C}_W \) with an Eastern culture and \( c_2 \in \mathcal{C}_E \), again using both implicit and explicit culture prompts. Specifically, we test four Western-Eastern cultural group pairs based on comparative findings from clinical psychology: Canadian and Chinese \cite{ryderCulturalShapingDepression2008}, Australian and Chinese \cite{parker2005depression}, American and Japanese \cite{arnaultSomaticDepressiveSymptoms2006}, and American and Indian \cite{biswasCrossCulturalVariations2016}. The template of \(\textsc{p}_{I} \) and \(\textsc{p}_{E} \) for Cultural Attribution task are provided below. 

\begin{tcolorbox}[
  colback=gray!10!white, 
  colframe=teal!80!black, 
  sharp corners=southwest, 
  title=\(\textsc{p}_{I} \), 
  coltitle=black, 
  fonttitle=\bfseries,
  listing only,
  listing options={
    basicstyle=\ttfamily,
    breaklines=true
  }
]
Analyze the given symptom and its description, and identify which demographic is most likely to express it. 

\vspace{1em}
Symptom: (symptom)

\vspace{1em}

Symptom’s description: (descriptions)

\vspace{1em}
Demographic list: [(country 1), (country 2)]
\end{tcolorbox}

\begin{tcolorbox}[
  colback=gray!10!white, 
  colframe=teal!80!black, 
  sharp corners=southwest, 
  title=\(\textsc{p}_{E} \), 
  coltitle=black, 
  fonttitle=\bfseries,
  listing only,
  listing options={
    basicstyle=\ttfamily,
    breaklines=true
  }
]
Analyze the given symptom and its description, and identify which demographic is most likely to express it. Choose one demographic from the provided list. Pay careful attention to cultural nuances and how different populations may exhibit or describe certain symptoms.

\vspace{1em}
Symptom: (symptom)

\vspace{1em}

Symptom’s description: (descriptions)

\vspace{1em}
Demographic list: [(country 1), (country 2)]
\end{tcolorbox}

Let \( P(c \mid \textsc{p}_{x}^{(c_1, c_2)}, s) \) denote the probability that the model selects cultural identity \( c \in \{c_1, c_2\} \) given symptom $s$ under prompt type $x$. We define the attribution bias as: 
\begin{align*}
P_{\Delta}(\textsc{p}_{x}^{(c_1, c_2)}, s) &= P(c_2 \mid \textsc{p}_{x}^{(c_1, c_2)}, s) \\
&\quad - P(c_1 \mid \textsc{p}_{x}^{(c_1, c_2)}, s)
\end{align*}
We then aggregate across symptom types: 
\begin{align*}
P_{\Delta}(\textsc{p}_{x}^{(c_1, c_2)}, \text{somatic})
&= \frac{1}{|\mathcal{S}_{\text{som}}|} \sum_{s \in S_{\text{som}}} \\
&\quad P_{\Delta}(\textsc{p}_{x}^{(c_1, c_2)}, s)
\end{align*}

\begin{align*}
P_{\Delta}(\textsc{p}_{x}^{(c_1, c_2)}, \text{psychological}) 
&= \frac{1}{|\mathcal{S}_{\text{psy}}|} \sum_{s \in S_{\text{psy}}} \\
&\quad P_{\Delta}(\textsc{p}_{x}^{(c_1, c_2)}, s)
\end{align*}

We define the cultural alignment measured in cultural attribution task as a vector of attribution biases, aggregated over somatic and psychological symptoms:
\begin{align*}
_2\mathcal{A}_x &= \big(P_{\Delta}(\textsc{p}_{x}^{(c_1, c_2)}, \text{somatic}), \\
&\quad P_{\Delta}(\textsc{p}_{x}^{(c_1, c_2)}, \text{psychological})\big)
\end{align*}

where the first component of \(_2\mathcal{A}_x\) captures the degree to which somatic symptoms are attributed to Eastern vs. Western cultural groups, while the second component captures the same for psychological symptoms.
Let \(_2\mathcal{A}_x^{\text{som}}\) and \(_2\mathcal{A}_x^{\text{psy}}\) denote the somatic and psychological components of \(_2\mathcal{A}_x\), respectively. 
Similar to symptom selection task, we denote $_2\mathcal{A}_x(l(c))$ as the cultural alignment tested by
LOC-P for Eastern country \( c_2 \in \mathcal{C}_E \), and $_2\mathcal{A}_x(Eng)$ as that by ENG-P. Under LOC-P condition, \(P_{\Delta}(\textsc{p}_{x}^{(c_1, c_2)}, s)\) is defined as,
\begin{align*}
P_{\Delta}(\textsc{p}_{x}^{(c_1, c_2)}, s, l(c_2)) 
&= P(c_2 \mid \textsc{p}_{x}^{(c_1, c_2)}, s, l(c_2)) \\
&\quad - P(c_1 \mid \textsc{p}_{x}^{(c_1, c_2)}, s, l(c_2))
\end{align*}

We test the hypotheses as: 

\smallskip

\noindent H1 is supported when $_2\mathcal{A}_x^{\text{som}} > 0, \; _2\mathcal{A}_x^{\text{psy}} < 0$, and \\H2 is supported when $_2\mathcal{A}_x^{\text{som}}(l(c)) > {_2}\mathcal{A}_x^{\text{som}}(Eng)$, \; $_2\mathcal{A}_x^{\text{psy}}(l(c)) < {_2}\mathcal{A}_x^{\text{psy}}(Eng)$.

\subsection{Results}
\subsubsection{Cultural Alignment in English Prompts (H1)}
\label{sec:H2-1}
In this task, we examine whether LLMs associate different depression symptoms with culturally appropriate groups.
Based on prior research, we expect LLMs to assign somatic symptoms \( S_{\text{som}} \) more frequently to Eastern cultural groups, and psychological symptoms \( S_{\text{psy}} \) to Western cultural groups. 

\begin{figure*}[h] 
    \centering
    \includegraphics[width=1.0\textwidth]{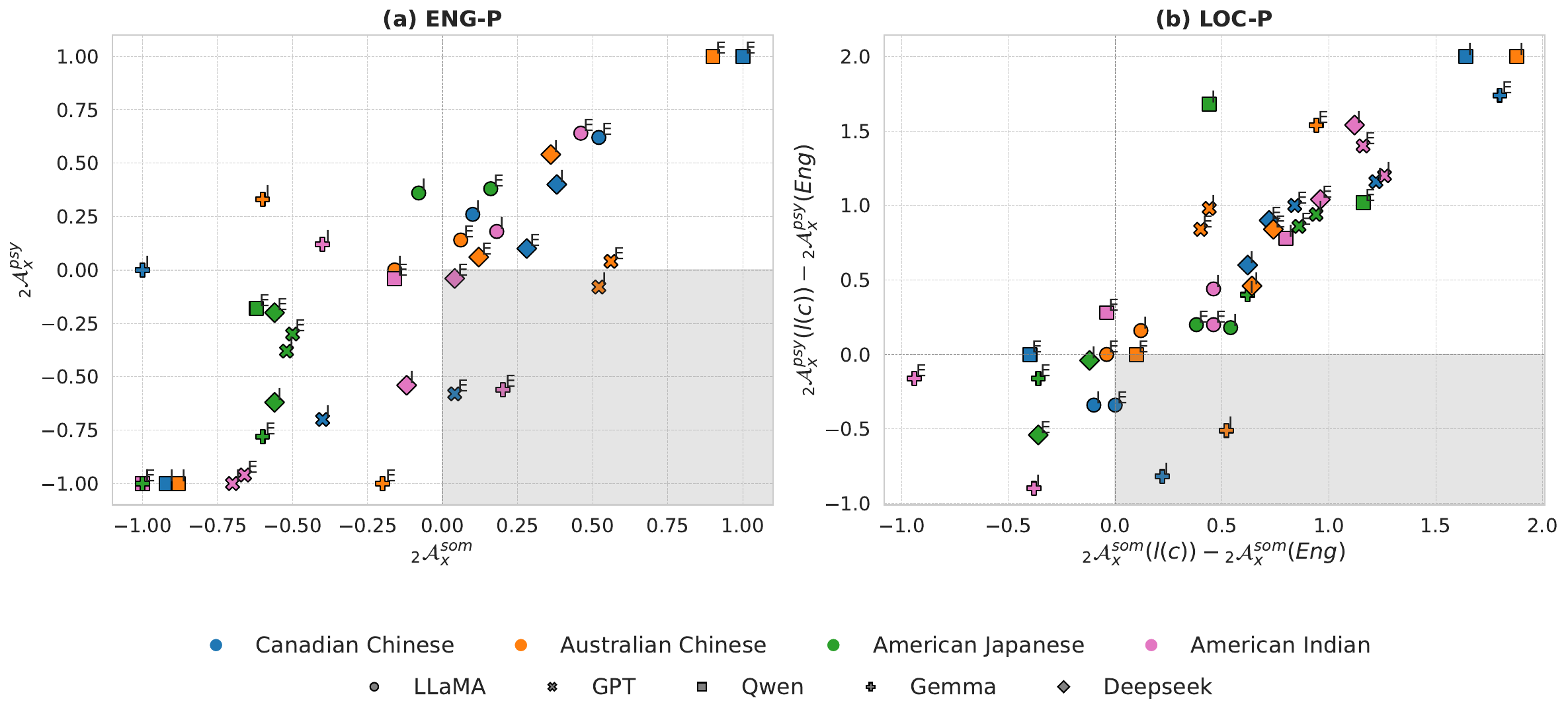} 
    \caption{Results on cultural attribution task. In (a), the $x$-axis shows somatic attribution $_2\mathcal{A}_x^{\text{som}}$, and the $y$-axis shows psychological attribution alignment $_2\mathcal{A}_x^{\text{psy}}$ under the ENG-P condition; values $>$ 0 on the $x$-axis and $<$ 0 on the $y$-axis indicate alignment with prior clinical psychology findings. In (b), the same region indicates  \emph{increased} alignment under the LOC-P condition. \emph{I} and \emph{E} indicate implicit (ICP, $\textsc{p}_{I}$) and explicit cultural prompt (ECP, $\textsc{p}_{E}$), respectively. The shaded quadrant represents culturally aligned attribution patterns to aid interpretation.}
    \label{fig:task2_alignment} 
\end{figure*}

Figure~\ref{fig:task2_alignment}(a) illustrates the alignment level of each model under the ENG-P condition. The $x$-axis represents the cultural alignment \(_2\mathcal{A}_x^{\text{som}}\), which is the average proportion of \( S_{\text{som}} \) assigned to Eastern groups minus that assigned to Western groups. 
The $y$-axis represents \(_2\mathcal{A}_x^{\text{psy}}\), the same calculation for \( S_{\text{psy}} \). 
Positive $x$-values and negative $y$-values indicate alignment with clinical expectations.

Overall, LLMs rarely exhibit this expected pattern. Only 4 out of 40 settings (5 models $\times$ 4 cultural group pairs $\times$ 2 prompt types) show alignment with prior findings: GPT for the Canadian–Chinese pair under \(\textsc{p}_{E} \), GPT for the Australian–Chinese pair under \(\textsc{p}_{I} \), Gemma and DeepSeek for the American–Indian pair under \(\textsc{p}_{E} \). 
In contrast, 31 settings show consistent attribution bias, with models favoring one cultural group (Eastern or Western) across both symptom types. 
These settings appear in quadrants where both $x$ and $y$ values are positive or both are negative, indicating that the same group is preferred regardless of symptom category.
We analyze this pattern in the next section. 
Alignment scores for all settings under the ENG-P condition are available in Table \ref{tab:symptom-region-model-all-en}.

In summary, \emph{H1 is not supported}, as most model behaviors under the ENG-P condition do not align with prior clinical psychology findings.

\begin{table*}[h]
\centering
\resizebox{\textwidth}{!}{%
\begin{tabular}{llcccccccccc}
\toprule
\textbf{Condition} & \textbf{Symptom Type}
& \multicolumn{2}{c}{\textbf{Llama}}
& \multicolumn{2}{c}{\textbf{GPT}}
& \multicolumn{2}{c}{\textbf{Qwen}}
& \multicolumn{2}{c}{\textbf{Gemma}}
& \multicolumn{2}{c}{\textbf{DeepSeek}} \\
&
& Western & Eastern
& Western & Eastern
& Western & Eastern
& Western & Eastern
& Western & Eastern \\
\midrule
Canadian Chinese - \(\textsc{p}_{I} \) & Somatic         & 0.45(0.14) & 0.55(0.14) & 0.70(0.32) & 0.30(0.32) & 0.96(0.09) & 0.04(0.09) & 1.00(0.00) & 0.00(0.00) & 0.31(0.16) & 0.69(0.16) \\
Canadian Chinese - \(\textsc{p}_{I} \) & Psychological   & 0.37(0.79) & 0.63(0.79) & 0.85(0.29) & 0.15(0.29) & 1.00(0.00) & 0.00(0.00) & 1.00(0.00) & 1.00(0.00) & 0.30(0.22) & 0.70(0.22) \\
Canadian Chinese - \(\textsc{p}_{E} \) & Somatic         & 0.24(0.15) & 0.76(0.15) & \textbf{0.48}(0.37) & \textbf{0.52}(0.37) & 0.00(0.00) & 1.00(0.00) & 1.00(0.00) & 0.00(0.00) & 0.36(0.09) & 0.64(0.09) \\
Canadian Chinese - \(\textsc{p}_{E} \) & Psychological   & 0.19(0.80) & 0.81(0.80) & \textbf{0.79}(0.32) & \textbf{0.21}(0.32) & 0.00(0.00) & 1.00(0.00) & 1.00(0.00) & 0.00(0.00) & 0.45(0.11) & 0.55(0.11) \\
Australian Chinese - \(\textsc{p}_{I} \) & Somatic       & 0.58(0.09) & 0.42(0.09) & \textbf{0.24}(0.43) & \textbf{0.76}(0.43) & 0.94(0.13) & 0.06(0.13) & 0.80(0.45) & 0.20(0.45) & 0.32(0.07) & 0.68(0.07) \\
Australian Chinese - \(\textsc{p}_{I} \) & Psychological & 0.50(0.04) & 0.50(0.04) & \textbf{0.54}(0.43) & \textbf{0.46}(0.43) & 1.00(0.00) & 0.00(0.00) & 0.33(0.50) & 0.66(0.50) & 0.23(0.21) & 0.77(0.21) \\
Australian Chinese - \(\textsc{p}_{E} \) & Somatic       & 0.47(0.11) & 0.53(0.11) & 0.22(0.44) & 0.78(0.44) & 0.05(0.12) & 0.95(0.12) & 0.60(0.55) & 0.40(0.55) & 0.44(0.12) & 0.56(0.12) \\
Australian Chinese - \(\textsc{p}_{E} \) & Psychological & 0.43(0.07) & 0.57(0.07) & 0.48(0.44) & 0.52(0.44) & 0.00(0.00) & 1.00(0.00) & 1.00(0.00) & 0.00(0.00) & 0.47(0.17) & 0.53(0.17) \\
American Japanese - \(\textsc{p}_{I} \) & Somatic        & 0.54(0.18) & 0.46(0.18) & 0.76(0.32) & 0.24(0.32) & 1.00(0.00) & 0.00(0.00) & 1.00(0.00) & 0.00(0.00) & 0.78(0.20) & 0.22(0.20) \\
American Japanese - \(\textsc{p}_{I} \) & Psychological  & 0.32(0.13) & 0.68(0.13) & 0.69(0.40) & 0.31(0.40) & 1.00(0.00) & 0.00(0.00) & 1.00(0.00) & 0.00(0.00) & 0.81(0.20) & 0.19(0.20) \\
American Japanese - \(\textsc{p}_{E} \) & Somatic        & 0.42(0.26) & 0.58(0.26) & 0.75(0.34) & 0.25(0.34) & 0.81(0.27) & 0.19(0.27) & 0.80(0.45) & 0.20(0.45) & 0.78(0.11) & 0.22(0.11) \\
American Japanese - \(\textsc{p}_{E} \) & Psychological  & 0.31(0.12) & 0.69(0.12) & 0.65(0.43) & 0.35(0.43) & 0.59(0.41) & 0.41(0.41) & 0.89(0.33) & 0.11(0.33) & 0.60(0.16) & 0.40(0.16) \\
American Indian - \(\textsc{p}_{I} \) & Somatic        & 0.41(0.21) & 0.59(0.21) & 0.85(0.33) & 0.15(0.33) & 1.00(0.00) & 0.00(0.00) & 0.60(0.55) & 0.20(0.45) & 0.56(0.32) & 0.44(0.32) \\
American Indian - \(\textsc{p}_{I} \) & Psychological  & 0.41(0.13) & 0.59(0.13) & 1.00(0.00) & 0.00(0.00) & 1.00(0.00) & 0.00(0.00) & 0.44(0.53) & 0.56(0.53) & 0.77(0.14) & 0.23(0.14) \\
American Indian - \(\textsc{p}_{E} \) & Somatic        & 0.27(0.11) & 0.73(0.11) & 0.83(0.36) & 0.17(0.36) & 0.58(0.44) & 0.42(0.44) & \textbf{0.40}(0.55) & \textbf{0.60}(0.55) & \textbf{0.48}(0.08) & \textbf{0.52}(0.08) \\
American Indian - \(\textsc{p}_{E} \) & Psychological  & 0.18(0.05) & 0.82(0.05) & 0.98(0.04) & 0.02(0.04) & 0.52(0.38) & 0.48(0.38) & \textbf{0.78}(0.44) & \textbf{0.22}(0.44) & \textbf{0.52}(0.08) & \textbf{0.48}(0.08) \\
\bottomrule
\end{tabular}
}
\caption{The proportions of selected Western or Eastern personas for given somatic or psychological symptoms across the experimental settings under the ENG-P condition. Experimental settings demonstrating cultural alignment are highlighted in bold.}
\label{tab:symptom-region-model-all-en}
\end{table*}

\subsubsection{Effect of Language on Attribution (H2)}
\label{sec:H2-2}
Figure \ref{fig:task2_alignment}(b) shows the change in cultural alignment for each experimental setting. Out of 40 settings, only two exhibit increased alignment, namely: Gemma Canadian-Chinese and Australian-Chinese pair under \(\textsc{p}_{I} \). As in the ENG-P condition, most models consistently favor one cultural group across symptom types.
Across the 40 settings, Eastern cultural groups are preferred in 27, Western groups in 10, and the remaining 3 show no consistent preference.  
Alignment scores and alignment improvement for all settings under the LOC-P condition are available in Table \ref{tab:symptom-region-model-eastern} and \ref{tab:model_culture_diffs}.

To statistically assess the impact of language on alignment, we conducted paired t-tests. 
An increase in alignment corresponds to a negative t-statistic for somatic symptoms and a positive one for psychological symptoms, as alignment improves when \(_2\mathcal{A}_x^{\text{som}}\) increases and \(_2\mathcal{A}_x^{\text{psy}}\) decreases.
However, as shown in Table \ref{tab:paired_ttest_task2_somatic} and \ref{tab:paired_ttest_task2_psy}, no model, cultural group, or prompt type meets this criterion. 
Overall, somatic symptoms tend to increase alignment, while psychological symptoms reduce it, indicating that Eastern groups are more likely to be associated with both symptom types under the LOC-P condition. 


Interestingly, alignment increases in symptom selection task under the LOC-P condition but decreases in cultural attribution task. This discrepancy likely stems from differences in task design and how language interacts with cultural framing. Symptom selection task evaluates whether models simulate culturally appropriate symptom expression when assigned a persona. In this context, using local language (LOC-P) reinforces cultural identity. For example, Eastern personas tend to select more somatic symptoms in LOC-P than in ENG-P, which enhances the contrast with Western personas. This reflects the model’s cultural reasoning based on persona and context. In contrast, cultural attribution task asks models to associate symptoms with cultural groups without assigning a persona. Since local language applies to both Western and Eastern cultural groups, the distinction between persona and linguistic framing becomes unclear. As a result, cultural attribution task captures how each language represents cultural identities rather than how models reason within them. The cultural attribution task results under LOC-P suggest that models may over-associate depression with Eastern groups, regardless of symptom type. This likely reflects biases in Eastern-language data, leading to a generalized ``depression equals Eastern" association rather than alignment with clinically grounded distinctions.

In summary, the \emph{results do not support H2} overall. 

\begin{table*}[h]
\centering
\resizebox{\textwidth}{!}{%
\begin{tabular}{llcccccccccc}
\toprule
\textbf{Condition} & \textbf{Symptom Type}
& \multicolumn{2}{c}{\textbf{Llama}}
& \multicolumn{2}{c}{\textbf{GPT}}
& \multicolumn{2}{c}{\textbf{Qwen}}
& \multicolumn{2}{c}{\textbf{Gemma}}
& \multicolumn{2}{c}{\textbf{DeepSeek}} \\
&
& Western & Eastern
& Western & Eastern
& Western & Eastern
& Western & Eastern
& Western & Eastern \\
\midrule
Canadian Chinese - \(\textsc{p}_{I} \) & Somatic         & 0.50(0.21) & 0.50(0.21) & 0.09(0.14) & 0.91(0.14) & 0.14(0.20) & 0.86(0.20) & 0.89(0.30) & 0.11(0.30) & 0.00(0.00) & 1.00(0.00) \\
Canadian Chinese - \(\textsc{p}_{I} \) & Psychological   & 0.54(0.10) & 0.46(0.10) & 0.27(0.36) & 0.73(0.36) & 0.00(0.00) & 1.00(0.00) & 0.91(0.16) & 0.09(0.16) & 0.00(0.00) & 1.00(0.00) \\
Canadian Chinese - \(\textsc{p}_{E} \) & Somatic         & 0.24(0.11) & 0.76(0.11) & 0.06(0.09) & 0.94(0.09) & 0.20(0.45) & 0.80(0.45) & 0.10(0.09) & 0.90(0.09) & 0.00(0.00) & 1.00(0.00) \\
Canadian Chinese - \(\textsc{p}_{E} \) & Psychological   & 0.36(0.10) & 0.64(0.10) & 0.29(0.38) & 0.71(0.38) & 0.00(0.00) & 1.00(0.00) & 0.13(0.09) & 0.87(0.09) & 0.00(0.00) & 1.00(0.00) \\
Australian Chinese - \(\textsc{p}_{I} \) & Somatic       & 0.52(0.09) & 0.48(0.09) & 0.02(0.02) & 0.98(0.02) & 0.00(0.00) & 1.00(0.00) & 0.54(0.14) & 0.46(0.14) & 0.00(0.00) & 1.00(0.00) \\
Australian Chinese - \(\textsc{p}_{I} \) & Psychological & 0.42(0.17) & 0.58(0.17) & 0.05(0.08) & 0.95(0.08) & 0.00(0.00) & 1.00(0.00) & 0.59(0.08) & 0.41(0.08) & 0.00(0.00) & 1.00(0.00) \\
Australian Chinese - \(\textsc{p}_{E} \) & Somatic       & 0.49(0.17) & 0.51(0.17) & 0.02(0.04) & 0.98(0.04) & 0.00(0.00) & 1.00(0.00) & 0.13(0.05) & 0.87(0.05) & 0.07(0.11) & 0.93(0.11) \\
Australian Chinese - \(\textsc{p}_{E} \) & Psychological & 0.43(0.14) & 0.57(0.14) & 0.06(0.13) & 0.94(0.13) & 0.00(0.00) & 1.00(0.00) & 0.23(0.07) & 0.77(0.07) & 0.05(0.09) & 0.95(0.09) \\
American Japanese - \(\textsc{p}_{I} \) & Somatic        & 0.27(0.19) & 0.73(0.19) & 0.29(0.40) & 0.71(0.40) & 0.78(0.31) & 0.22(0.31) & 0.69(0.11) & 0.31(0.11) & 0.84(0.15) & 0.16(0.15) \\
American Japanese - \(\textsc{p}_{I} \) & Psychological  & 0.23(0.13) & 0.77(0.13) & 0.22(0.42) & 0.78(0.42) & 0.16(0.25) & 0.84(0.25) & 0.80(0.09) & 0.20(0.09) & 0.83(0.08) & 0.17(0.08) \\
American Japanese - \(\textsc{p}_{E} \) & Somatic        & 0.28(0.05) & 0.82(0.05) & 0.32(0.46) & 0.68(0.46) & 0.23(0.37) & 0.77(0.37) & 0.98(0.01) & 0.02(0.01) & 0.96(0.05) & 0.04(0.05) \\
American Japanese - \(\textsc{p}_{E} \) & Psychological  & 0.23(0.14) & 0.81(0.14) & 0.22(0.44) & 0.78(0.44) & 0.08(0.23) & 0.92(0.23) & 0.97(0.02) & 0.03(0.02) & 0.87(0.07) & 0.13(0.07) \\
American Indian - \(\textsc{p}_{I} \) & Somatic        & 0.18(0.09) & 0.82(0.09) & 0.22(0.25) & 0.78(0.25) & 0.60(0.55) & 0.40(0.55) & 0.89(0.07) & 0.11(0.07) & 0.00(0.00) & 1.00(0.00) \\
American Indian - \(\textsc{p}_{I} \) & Psychological  & 0.19(0.06) & 0.81(0.06) & 0.40(0.33) & 0.60(0.33) & 0.61(0.49) & 0.39(0.49) & 0.89(0.04) & 0.11(0.04) & 0.00(0.00) & 1.00(0.00) \\
American Indian - \(\textsc{p}_{E} \) & Somatic        & 0.04(0.02) & 0.96(0.02) & 0.25(0.29) & 0.75(0.29) & 0.60(0.55) & 0.40(0.55) & 0.87(0.03) & 0.13(0.03) & 0.00(0.00) & 1.00(0.00) \\
American Indian - \(\textsc{p}_{E} \) & Psychological  & 0.08(0.05) & 0.92(0.05) & 0.28(0.22) & 0.72(0.22) & 0.38(0.42) & 0.62(0.42) & 0.86(0.10) & 0.14(0.10) & 0.00(0.00) & 1.00(0.00) \\
\bottomrule
\end{tabular}
}
\caption{The proportions of selected Western or Eastern personas for given somatic or psychological symptoms across the experimental settings under the LOC-P condition.}
\label{tab:symptom-region-model-eastern}
\end{table*}

\begin{table}[t]
\centering
\resizebox{\columnwidth}{!}{%
\begin{tabular}{llccccc}
\toprule
\textbf{Condition} & \textbf{Symptom Type} & \textbf{Llama} & \textbf{GPT} & \textbf{Qwen} & \textbf{Gemma} & \textbf{DeepSeek} \\
\midrule
Canadian Chinese - I & Somatic        & -0.10 & 1.22 & 1.64 & \textbf{0.22} & 0.62 \\
Canadian Chinese - I & Psychological  & -0.34 & 1.16 & 2.00 & \textbf{-0.82} & 0.60 \\
Canadian Chinese - E & Somatic        &  0.00 & 0.84 & -0.40 & 1.80 & 0.72 \\
Canadian Chinese - E & Psychological  & -0.34 & 1.00 & 0.00 & 1.74 & 0.90 \\
Australian Chinese - I & Somatic      &  0.12 & 0.44 & 1.88 & \textbf{0.52} & 0.64 \\
Australian Chinese - I & Psychological&  0.16 & 0.98 & 2.00 & \textbf{-0.51} & 0.46 \\
Australian Chinese - E & Somatic      & -0.04 & 0.40 & 0.10 & 0.94 & 0.74 \\
Australian Chinese - E & Psychological&  0.00 & 0.84 & 0.00 & 1.54 & 0.84 \\
American Japanese - I & Somatic       &  0.54 & 0.94 & 0.44 & 0.62 & -0.12 \\
American Japanese - I & Psychological &  0.18 & 0.94 & 1.68 & 0.40 & -0.04 \\
American Japanese - E & Somatic       &  0.38 & 0.86 & 1.16 & -0.36 & -0.36 \\
American Japanese - E & Psychological &  0.20 & 0.86 & 1.02 & -0.16 & -0.54 \\
American Indian - I & Somatic        &  0.46 & 1.26 & 0.80 & -0.38 & 1.12 \\
American Indian - I & Psychological  &  0.44 & 1.20 & 0.78 & -0.90 & 1.54 \\
American Indian - E & Somatic        &  0.46 & 1.16 & -0.04 & -0.94 & 0.96 \\
American Indian - E & Psychological  &  0.20 & 1.40 & 0.28 & -0.16 & 1.04 \\
\bottomrule
\end{tabular}
}
\caption{$_2\mathcal{A}_x^{\text{som}}(l(c)) - {_2}\mathcal{A}_x^{\text{som}}(Eng)$ and $_2\mathcal{A}_x^{\text{psy}}(l(c)) - {_2}\mathcal{A}_x^{\text{psy}}(Eng)$ for each experimental setting. \emph{Somatic} indicates $_2\mathcal{A}_x^{\text{som}}(l(c)) - {_2}\mathcal{A}_x^{\text{som}}(Eng)$ and \emph{Psychological} indicates $_2\mathcal{A}_x^{\text{psy}}(l(c)) - {_2}\mathcal{A}_x^{\text{psy}}(Eng)$. Experimental settings demonstrating the increased cultural alignment are highlighted in bold.}
\label{tab:model_culture_diffs}
\end{table}

\begin{table*}[h]
    \centering
    \resizebox{\textwidth}{!}{%
    \begin{tabular}{lrrrrrrrrrrrr}
        \toprule
        \textbf{Somatic} & \textbf{All} & \textbf{Llama} & \textbf{GPT} & \textbf{Qwen} & \textbf{Gemma} & \textbf{DeepSeek} & \textbf{Ca-Ch} & \textbf{Au-Ch} & \textbf{Am-Ja} & \textbf{Am-In} & \textbf{\(\textsc{p}_{I} \)} & \textbf{\(\textsc{p}_{E} \)} \\
        \midrule
        t-stat & \textbf{-5.34} & \textbf{-2.48} & \textbf{-7.60} & \textbf{-2.41} & -0.99 & \textbf{-2.97} & \textbf{-2.80} & \textbf{-3.30} & \textbf{-2.42} & -2.10 & \textbf{-4.98} & \textbf{-2.79}\\
        p-value & \textbf{0.00} & \textbf{0.04} & \textbf{0.00} & \textbf{0.05} & 0.36 & \textbf{0.02} & \textbf{0.02} & \textbf{0.01} & \textbf{0.04} & 0.07 & \textbf{0.00} & \textbf{0.01}\\
        \bottomrule
    \end{tabular}
    }
    \caption{Paired t-test by models, cultural group pairs, and prompt types for \textbf{somatic symptoms}. Statistically significant values (p $<$ 0.05) are bolded.}
    \label{tab:paired_ttest_task2_somatic}
\end{table*}

\begin{table*}[h]
    \centering
    \resizebox{\textwidth}{!}{%
    \begin{tabular}{lrrrrrrrrrrrr}
        \toprule
        \textbf{Psychological} & \textbf{All} & \textbf{Llama} & \textbf{GPT} & \textbf{Qwen} & \textbf{Gemma} & \textbf{DeepSeek} & \textbf{Ca-Ch} & \textbf{Au-Ch} & \textbf{Am-Ja} & \textbf{Am-In} & \textbf{\(\textsc{p}_{I} \)} & \textbf{\(\textsc{p}_{E} \)} \\
        \midrule
        t-stat & \textbf{-4.69} & -0.64 & \textbf{-15.47} & \textbf{-3.24} & -0.39 & \textbf{-2.61} & -1.98 & \textbf{-2.61} & -2.16 & \textbf{-2.42} & \textbf{-3.08} & \textbf{-3.59}\\
        p-value & \textbf{0.00} & 0.54 & \textbf{0.00} & \textbf{0.01} & 0.71 & \textbf{0.03} & 0.09 & \textbf{0.03} & 0.06 & \textbf{0.04} & \textbf{0.01} & \textbf{0.00}\\
        \bottomrule
    \end{tabular}
    }
    \caption{Paired t-test by models, cultural group pairs, and prompt types for \textbf{psychological symptoms}. Statistically significant values (p $<$ 0.05) are bolded.}
    \label{tab:paired_ttest_task2_psy}
\end{table*}

\subsubsection{Individual Symptom Level Analysis}
Similar to symptom selection task, we also conducted symptom level analysis for psychological symptoms in cultural attribution task. We examined whether LLMs exhibited similar patterns of statistically significant differences (p $<$ 0.05) in cultural attributions across Eastern and Western cultural personas.

\textbf{Australian-Chinese Pair.} For ``Depressed mood'', six conditions showed alignment: (Deepseek, LOC-P), (Gemma, ENG-P), (Gemma, LOC-P), (GPT, ENG-P), (GPT, LOC-P), and (Qwen, ENG-P). For ``Loss of interest'', three conditions aligned: (Gemma, ENG-P), (Gemma, LOC-P), and (Qwen, ENG-P). For ``Suicidal thoughts'', alignment was observed in three conditions: (Llama, ENG-P), (Llama, LOC-P), and (Qwen, LOC-P).

\textbf{Canadian-Chinese Pair.} We found alignment in four conditions for ``Depressed mood'': (Deepseek, ENG-P), (Llama, ENG-P), (Llama, LOC-P), and (Qwen, LOC-P).

\textbf{American-Indian Pair.} For ``Decreased interest in pleasurable activities'', four conditions aligned: (Deepseek, ENG-P), (GPT, ENG-P), (GPT, LOC-P), and (Qwen, ENG-P). For ``Pessimistic view of the future'', alignment was observed in seven conditions: (Deepseek, ENG-P), (Deepseek, LOC-P), (Gemma, LOC-P), (GPT, ENG-P), (GPT, LOC-P), (Qwen, ENG-P), and (Qwen, LOC-P). For ``Ideas of self-harm or suicide'', nine conditions showed alignment: (Deepseek, ENG-P), (Deepseek, LOC-P), (Gemma, ENG-P), (Gemma, LOC-P), (GPT, ENG-P), (GPT, LOC-P), (Llama, ENG-P), (Qwen, ENG-P), and (Qwen, LOC-P).

Out of 10 total experimental coditions for each symptom (2 prompt types × 5 models), the number of aligned conditions ranges from three to nine, with ``Ideas of self-harm or suicide'' showing the highest number of alignments. While cultural attribution task demonstrates overall improved alignment compared to symptom selection task, this improvement is likely attributable to the task design, which explicitly required LLMs to choose between a Western or Eastern country. It may also reflect consistent attribution biases, as discussed in Sections~\ref{sec:H2-1} and~\ref{sec:H2-2}.

\subsubsection{Model Preferences for Eastern vs. Western Cultural Groups}
\label{sec:modelpreferencemain}
To further explore cultural attribution patterns in LLMs, we analyze each model's overall tendency to favor either Eastern or Western cultural groups when assigning depression symptoms. We calculate the averaged \(P_{\Delta}(\textsc{p}_{x}^{(c_1, c_2)}, s)\) across the symptoms. 

Table \ref{tab:average_country} shows that GPT, Qwen, and Gemma predominantly assign symptoms to Western cultural groups in 6 out of 8 country pairs, indicating a stronger association between depression and Western identities. This may reflect the overrepresentation of Western cultural perspectives in their English-language training data. DeepSeek shows a possible group-level cultural bias: it favors Eastern groups for the Australian–Chinese and Canadian–Chinese pairs, but favors Western groups for the American–Japanese and American–Indian pairs. 
In contrast, Llama consistently assigns symptoms to Eastern groups in 7 out of 8 country pairs, suggesting a potential bias toward Eastern cultural associations with depression. 
More refined methods may be needed to determine whether these biases are due to training data distributions or other model-internal factors. 

Prompt type also influences the direction of cultural bias. Llama, GPT, and Qwen all exhibit an increased Eastern preference under \(\textsc{p}_{E} \). Qwen shows the most dramatic change, particularly for the Australian–Chinese and Canadian–Chinese pairs, changing from strong Western to strong Eastern preference. DeepSeek shows mixed results. It starts favoring Western personas with \(\textsc{p}_{E} \) for the Australian–Chinese and Canadian–Chinese pairs, whereas it starts favoring Eastern personas for the American–Japanese and American–Indian pairs, indicating a culturally contingent response pattern. These findings highlight that LLMs' cultural attributions are not fixed but can be modulated by contextual cues embedded in prompts.

Under the LOC-P condition, Llama and Gemma show results consistent with those observed under the ENG-P condition: Llama predominantly exhibits an Eastern persona bias, while Gemma demonstrates a Western persona bias (Table~\ref{tab:average_country_zh}). In contrast, GPT and Qwen both shifted to an Eastern persona bias under LOC-P. DeepSeek also reversed its bias for the American-Indian pair, changing from a Western to an Eastern persona bias.

\begin{table}[h]
    \centering
    \resizebox{\columnwidth}{!}{%
    \begin{tabular}{lrrrrr}
        \toprule
        & \textbf{Llama} & \textbf{GPT} & \textbf{Qwen} & \textbf{Gemma} & \textbf{DeepSeek} \\
        \midrule
        Au-Ch \(\textsc{p}_{I} \) & 0.21(0.21) & -0.59(0.59) & -0.97(0.10) & -1.00(0.00) & 0.34(0.34) \\
        Au-Ch \(\textsc{p}_{E} \) & 0.58(0.21) & -0.35(0.72) & 1.00(0.00) & -1.00(0.00) & 0.17(0.22) \\
        Ca-Ch \(\textsc{p}_{I} \) & -0.06(0.15) & 0.13(0.88) & -0.96(0.15) & 0.00(1.04) & 0.48(0.35) \\
        Ca-Ch \(\textsc{p}_{E} \) & 0.11(0.17) & 0.23(0.88) & 0.96(0.14) & -0.71(0.73) & 0.08(0.31) \\
        Am-Ja \(\textsc{p}_{I} \) & 0.21(0.36) & -0.43(0.73) & -1.00(0.00) & -1.00(0.00) & -0.60(0.39) \\
        Am-Ja \(\textsc{p}_{E} \) & 0.36(0.36) & -0.36(0.78) & -0.33(0.75) & -0.71(0.73) & -0.33(0.34) \\
        Am-In \(\textsc{p}_{I} \) & 0.18(0.31) & -0.89(0.40) & -1.00(0.00) & 0.08(1.00) & -0.40(0.47) \\
        Am-In \(\textsc{p}_{E} \) & 0.58(0.17) & -0.85(0.43) & -0.08(0.77) & -0.29(0.99) & -0.01(0.16) \\
        \bottomrule
    \end{tabular}
    }
    \caption{\(P_{\Delta}(\textsc{p}_{x}^{(c_1, c_2)}, s)\) is averaged across the symptoms under the ENG-P condition. Values in parentheses indicate  standard deviations.}
    \label{tab:average_country}
\end{table}

\begin{table}[h]
    \centering
    \resizebox{\columnwidth}{!}{%
    \begin{tabular}{lrrrrr}
        \toprule
        & \textbf{Llama} & \textbf{GPT} & \textbf{Qwen} & \textbf{Gemma} & \textbf{DeepSeek} \\
        \midrule
        Au-Ch \(\textsc{p}_{I} \) & -0.14(0.30) & 0.92(0.13) & 1.00(0.00) & -0.81(0.20) & 1.00(0.00) \\
        Au-Ch \(\textsc{p}_{E} \) & 0.08(0.29) & 0.91(0.21) & 1.00(0.00) & 0.75(0.12) & 0.88(0.19) \\
        Ca-Ch \(\textsc{p}_{I} \) & -0.05(0.28) & 0.58(0.61) & 0.90(0.26) & -0.14(0.42) & 1.00(0.00) \\
        Ca-Ch \(\textsc{p}_{E} \) & 0.37(0.24) & 0.59(0.65) & 0.86(0.53) & 0.62(0.19) & 1.00(0.00) \\
        Am-Ja \(\textsc{p}_{I} \) & 0.51(0.30) & 0.51(0.81) & 0.23(0.80) & -0.51(0.22) & -0.67(0.21) \\
        Am-Ja \(\textsc{p}_{E} \) & 0.51(0.23) & 0.49(0.86) & 0.74(0.56) & -0.95(0.03) & -0.81(0.15) \\
        Am-In \(\textsc{p}_{I} \) & 0.62(0.14) & 0.32(0.62) & -0.21(0.97) & -0.78(0.10) & 0.99(0.01) \\
        Am-In \(\textsc{p}_{E} \) & 0.87(0.09) & 0.45(0.47) & 0.09(0.92) & -0.73(0.16) & 1.00(0.00) \\
        \bottomrule
    \end{tabular}
    }
    \caption{\(P_{\Delta}(\textsc{p}_{x}^{(c_1, c_2)}, s)\) is averaged across the symptoms under the LOC-P condition. Values in parentheses indicate  standard deviations.}
    \label{tab:average_country_zh}
\end{table}

\subsubsection{Determinism in Cultural Attribution}
Similar to symptom selection task, we compute the Gini coefficient for each experimental setting under the ENG-P condition using the attribution bias \(P_{\Delta}(\textsc{p}_{x}^{(c_1, c_2)}, s)\) (see Table \ref{tab:gini_task2}). Lower Gini values indicate a consistent level of preference for one cultural group across symptoms, while higher values reflect greater variation of preference level in attribution across symptoms.

Consistent with the determinism analysis in symptom selection task, results from cultural attribution task show that Qwen and Gemma tend to exhibit the highest deterministic behaviors (4 and 5 settings respectively). This indicates a persistent level of the attribution bias across symptoms, regardless of symptom category. In contrast, the model with the highest Gini coefficient varies depending on the experimental setting. Under the LOC-P condition, unlike the ENG-P condition, DeepSeek consistently shows the lower Gini coefficient values. The model with the highest Gini coefficient is not consistent across the experimental settings (Table \ref{tab:gini_task2_zh}).

\begin{table}[h]
    \centering
    \resizebox{\columnwidth}{!}{%
    \begin{tabular}{lrrrrr}
        \toprule
        & \textbf{Llama} & \textbf{GPT} & \textbf{Qwen} & \textbf{Gemma} & \textbf{DeepSeek} \\
        \midrule
        Au-Ch \(\textsc{p}_{I} \) & 1.31 & \textbf{3.50} & 0.04 & \underline{0.00} & 0.40 \\
        Au-Ch \(\textsc{p}_{E} \) & 0.86 & 2.00 & \underline{0.04} & 0.34 & \textbf{2.18} \\
        Ca-Ch \(\textsc{p}_{I} \) & 0.51 & 0.47 & 0.03 & \underline{0.00} & \textbf{0.54} \\
        Ca-Ch \(\textsc{p}_{E} \) & 0.17 & \textbf{1.10} & \underline{0.00} & \underline{0.00} & 0.65 \\
        Am-Ja \(\textsc{p}_{I} \) & \textbf{0.88} & 0.85 & \underline{0.00} & \underline{0.00} & 0.33 \\
        Am-Ja \(\textsc{p}_{E} \) & 0.57 & 1.09 & \textbf{1.16} & \underline{0.34} & 0.54 \\
        Am-In \(\textsc{p}_{I} \) & 0.95 & 0.11 & \underline{0.00} & \textbf{6.93} & 0.59 \\
        Am-In \(\textsc{p}_{E} \) & \underline{0.14} & 0.15 & 5.00 & 1.60 & \textbf{15.80} \\
        Average  & \underline{0.67} & 1.16 & 0.78 & 1.15 & \textbf{2.63} \\
        \bottomrule
    \end{tabular}
    }
    \caption{The Gini coefficient for
each experimental setting under the ENG-P condition. The highest values are highlighted in bold, while the lowest are underlined.}
    \label{tab:gini_task2}
\end{table}

\begin{table}[h]
    \centering
    \resizebox{\columnwidth}{!}{%
    \begin{tabular}{lrrrrr}
        \toprule
        & \textbf{Llama} & \textbf{GPT} & \textbf{Qwen} & \textbf{Gemma} & \textbf{DeepSeek} \\
        \midrule
        Au-Ch \(\textsc{p}_{I} \) & \textbf{1.10} & 0.06 & \underline{0.00} & 0.12 & \underline{0.00} \\
        Au-Ch \(\textsc{p}_{E} \) & \textbf{2.10} & 0.08 & \underline{0.00} & 0.07 & 0.10 \\
        Ca-Ch \(\textsc{p}_{I} \) & \textbf{2.78} & 0.48 & 0.10 & 1.51 & \underline{0.00} \\
        Ca-Ch \(\textsc{p}_{E} \) & 0.36 & \textbf{0.50} & 0.15 & 0.17 & \underline{0.00} \\
        Am-Ja \(\textsc{p}_{I} \) & 0.31 & 0.70 & \textbf{1.80} & 0.23 & \underline{0.18} \\
        Am-Ja \(\textsc{p}_{E} \) & 0.25 & \textbf{0.78} & 0.29 & \underline{0.02} & 0.10 \\
        Am-In \(\textsc{p}_{I} \) & 0.12 & 1.03 & \textbf{2.21} & 0.07 & \underline{0.00} \\
        Am-In \(\textsc{p}_{E} \) & 0.05 & 0.56 & \textbf{5.59} & 0.10 & \underline{0.00} \\
        \bottomrule
    \end{tabular}
    }
    \caption{The Gini coefficient for each experimental setting under the LOC-P condition. The highest values are highlighted in
bold, while the lowest are underlined.}
    \label{tab:gini_task2_zh}
\end{table}

\subsubsection{Sensitivity to Cultural Group Pairs}
Similar to symptom selection task, we assess each model's sensitivity to different cultural group pairs and prompts. \emph{Cultural group pair sensitivity} measures how well a model differentiates between cultural group pairs, calculated as the average cosine similarity of attribution bias  \(P_{\Delta}(\textsc{p}_{x}^{(c_1, c_2)}, s)\) across all group pairs within the same prompt type. Lower cosine similarity values indicate higher sensitivity.

Under the ENG-P condition, Table \ref{tab:sensitivity-task2} shows that Qwen exhibits the highest cultural group pair sensitivity with \(\textsc{p}_{E} \) (cosine similarity = 0.10) and the extreme change between prompt type (0.99 $\rightarrow$ 0.10), suggesting explicit instructions significantly enhance Qwen's differentiation across group pairs. In contrast, Llama, Gemma, and DeepSeek show reduced sensitivity from \(\textsc{p}_{I} \) to \(\textsc{p}_{E} \), indicating less benefit from explicit prompting.
Overall, DeepSeek displays the strongest sensitivity to cultural group pairs (0.01 with \(\textsc{p}_{I} \), 0.24 with \(\textsc{p}_{E} \)), suggesting highly responsive to cultural cues provided in the prompts.

Under the LOC-P condition, DeepSeek maintains relatively higher sensitivity for both \(\textsc{p}_{I} \) and \(\textsc{p}_{E} \) prompts, reinforcing its responsiveness to cultural variation. Llama and Gemma show greater sensitivity for \(\textsc{p}_{I} \) and \(\textsc{p}_{E} \) prompts respectively. 

\begin{table*}[ht!]
\centering
\resizebox{\textwidth}{!}{%
\begin{tabular}{ccccccccccc}
\toprule
\multicolumn{2}{c}{Llama} & \multicolumn{2}{c}{GPT} & \multicolumn{2}{c}{Qwen} & \multicolumn{2}{c}{Gemma} & \multicolumn{2}{c}{DeepSeek} \\
ENG-P & LOC-P & ENG-P & LOC-P & ENG-P & LOC-P & ENG-P & LOC-P & ENG-P & LOC-P \\
\midrule
0.38/0.81 & -0.02/0.65 & 0.49/0.43 & 0.54/0.60 & 0.99/0.10 & 0.23/0.52 & 0.23/0.50 & 0.61/-0.32 & 0.01/0.24 & 0.02/0.01 \\
\bottomrule
\end{tabular}
}
\caption{Average cosine similarities across cultural group pairs under the ENG-P and LOC-P condition. Smaller cosine similarity indicates more sensitivity. Two cosine similarity values in each prompt type correspond to the cosine values for \(\textsc{p}_{I} \) and \(\textsc{p}_{E} \). }
\label{tab:sensitivity-task2}
\end{table*}

\subsubsection{The Overall Results}
\begin{figure*}[h]
  \centering
  \includegraphics[width=0.9\textwidth]{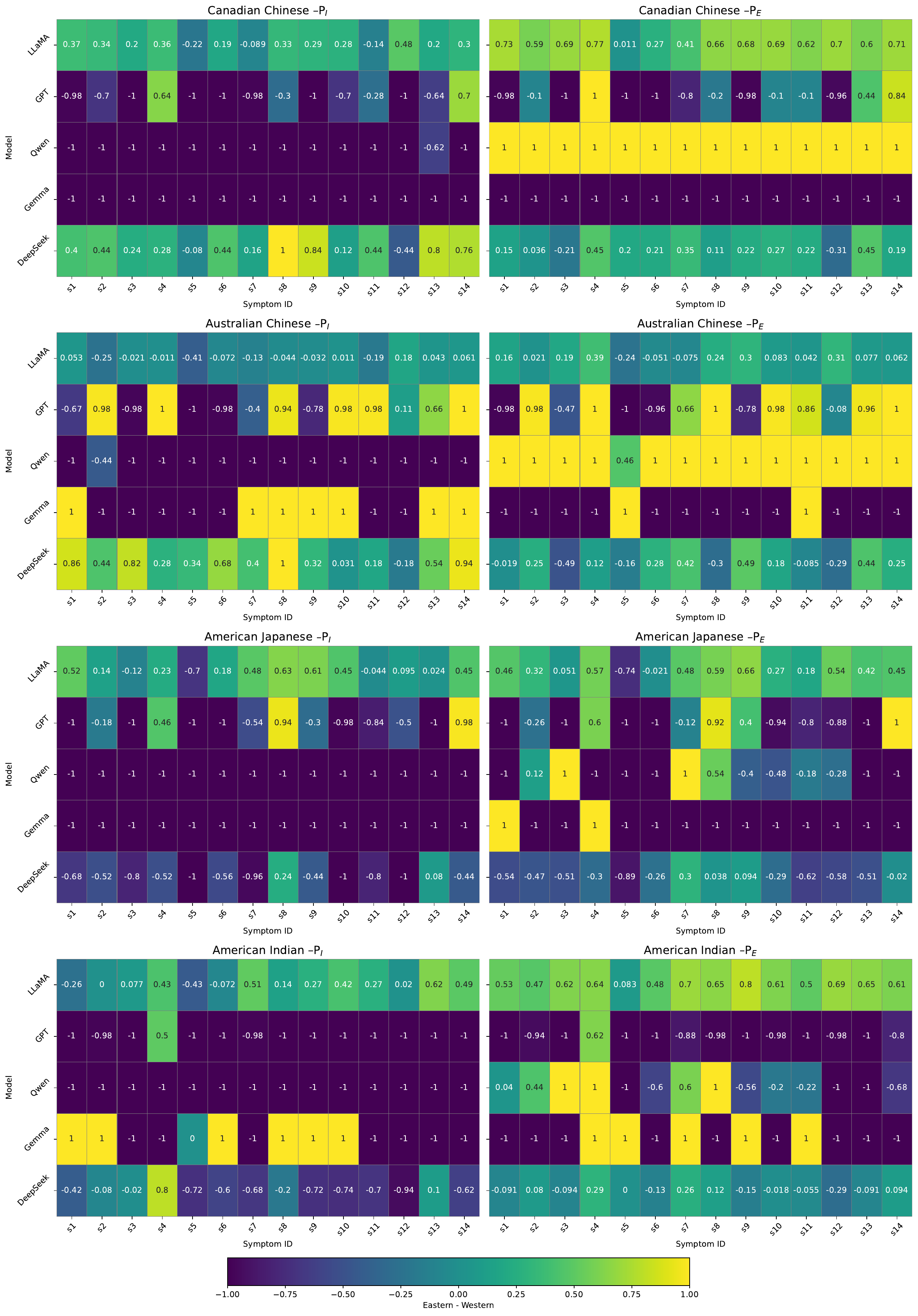}
  \caption{\(P_{\Delta}(\textsc{p}_{x}^{(c_1, c_2)}, s)\) across models for four cultural group pairs under the ENG-P condition.}
  \label{fig:heatmap-t2}
\end{figure*}

\begin{figure*}[h]
  \centering
  \includegraphics[width=0.9\textwidth]{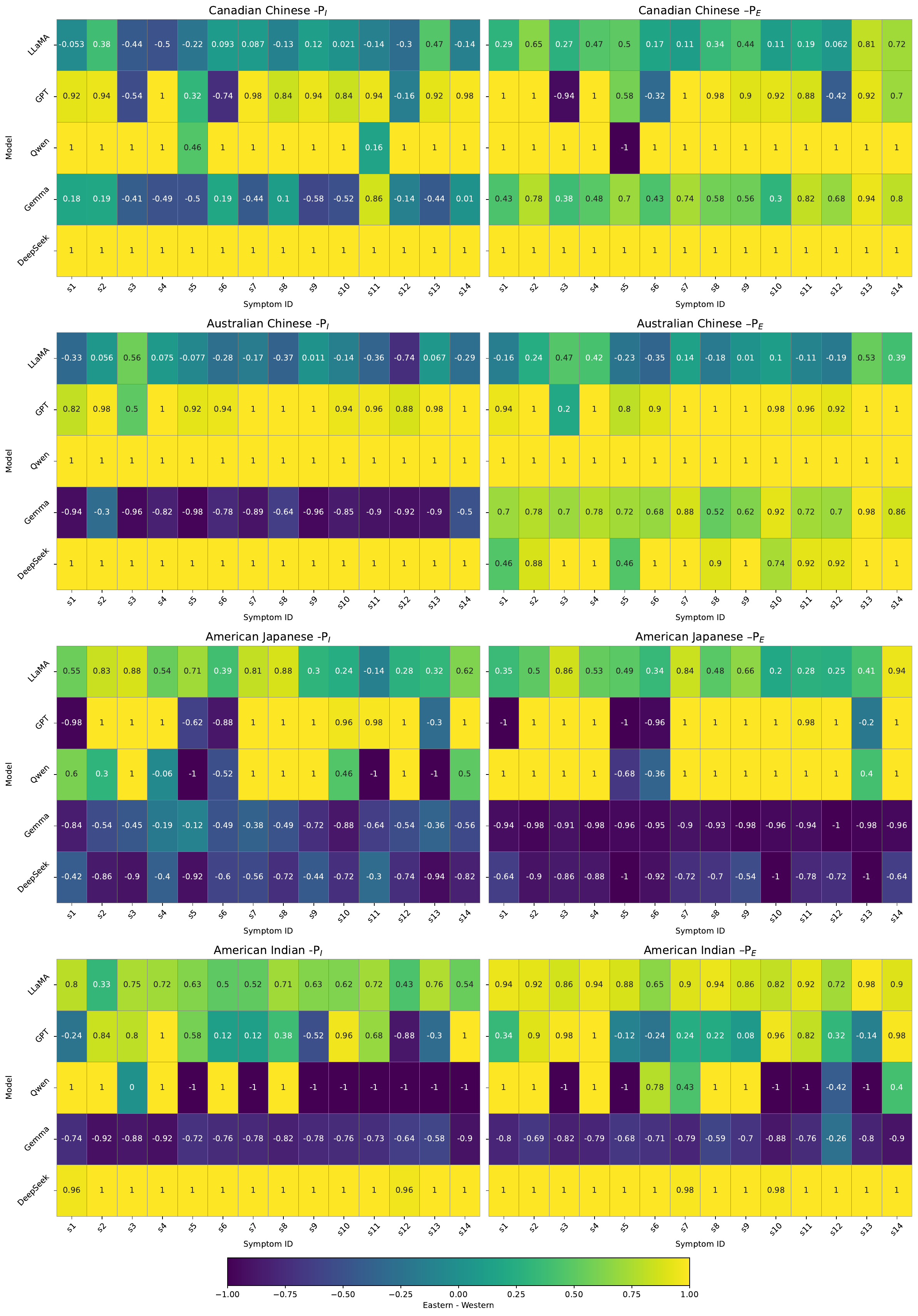}
  \caption{\(P_{\Delta}(\textsc{p}_{x}^{(c_1, c_2)}, s)\) across models for four cultural group pairs under the LOC-P condition.}
  \label{fig:heatmap-t2-zh}
\end{figure*}

Figure \ref{fig:heatmap-t2} (under the ENG-P condition) and  \ref{fig:heatmap-t2-zh} (under the LOC-P condition) display the proportions of selected Eastern personas minus that of Western personas (\(P_{\Delta}(\textsc{p}_{x}^{(c_1, c_2)}, s)\)) across all the Task2 experimental settings. The positive value indicates that Eastern persona was more likely to be selected, while negative value indicate the opposite. 

\end{document}